\documentclass[lettersize,journal]{IEEEtran}
\usepackage{algorithmic}
\usepackage{array}
\usepackage[caption=false,font=footnotesize,labelfont=sf]{subfig}
\usepackage{textcomp}
\usepackage{stfloats}
\usepackage{verbatim}
\def\BibTeX{{\rm B\kern-.05em{\sc i\kern-.025em b}\kern-.08em
    T\kern-.1667em\lower.7ex\hbox{E}\kern-.125emX}}
\usepackage{balance}

\usepackage[utf8]{inputenc} 
\usepackage[T1]{fontenc}    
\usepackage{hyperref}       
\usepackage{url}            
\usepackage{booktabs}       
\usepackage{amsfonts}       
\usepackage{nicefrac}       
\usepackage{microtype}      
\usepackage[table]{xcolor}  
\usepackage{colortbl}
\usepackage{graphicx}
\usepackage{amsmath}
\usepackage{mathtools}
\usepackage{amsthm}
\usepackage{bm}
\usepackage{bbm}
\usepackage{caption}
\usepackage{multirow}
\usepackage{makecell}
\usepackage{soul}
\usepackage{comment}
\usepackage{tablefootnote}
\usepackage{adjustbox}
\usepackage{svg}
\usepackage{tabularx}
\usepackage{wasysym}
\usepackage{todonotes}
\usepackage{tikz}
\usepackage{cite}
\usetikzlibrary{positioning, shadows, shapes.geometric, calc}

\begin{document}
\title{A Survey on Decentralized Federated Learning}
\author{Edoardo Gabrielli, Anthony Di Pietro, Dario Fenoglio, Giovanni Pica, Gabriele Tolomei
\thanks{
    Edoardo Gabrielli (corresponding author) is with the Department of Computer, Control and Management Engineering, Sapienza University of Rome, Italy. E-mail: edoardo.gabrielli@uniroma1.it.
    
    Anthony Di Pietro, Giovanni Pica, and Gabriele Tolomei are with the Department of Computer Science, Sapienza University of Rome, Italy.

    Dario Fenoglio is with the Università della Svizzera Italiana, Lugano, Switzerland.
    }
}

\markboth{Journal of \LaTeX\ Class Files,~Vol.~18, No.~9, September~2020}%
{How to Use the IEEEtran \LaTeX \ Templates}

\maketitle

\begin{abstract}
Federated learning (FL) enables collaborative training without pooling raw data, but standard FL relies on a central coordinator, which introduces a single point of failure and concentrates trust in the orchestration infrastructure. 
Decentralized federated learning (DFL) removes the coordinator and replaces client–server orchestration with peer-to-peer coordination, making learning dynamics topology-dependent and reshaping the associated security, privacy, and systems trade-offs. 
This survey systematically reviews DFL methods from 2018 through early 2026 and organizes them into two architectural families: traditional distributed FL and blockchain-based FL. 
We then propose a unified, challenge-driven taxonomy that maps both families to the core bottlenecks they primarily address, and we summarize prevailing evaluation practices and their limitations, exposing gaps in the literature. 
Finally, we distill lessons learned and outline research directions, emphasizing topology-aware threat models, privacy notions that reflect decentralized exposure, incentive mechanisms robust to manipulation, and the need to explicitly define whether the objective is a single global model or personalized solutions in decentralized settings.
\end{abstract}


\newcommand{\Prob}{\mathbb{P}}
\newcommand{\R}{\mathbb{R}}
\newcommand{\Z}{\mathbb{Z}}
\newcommand{\E}{\mathbb{E}}
\newcommand{\insta}{\bm{x}}
\newcommand{\X}{X}
\newcommand{\dataset}{\mathcal{D}}
\newcommand{\train}{\dataset_{\text{train}}}
\newcommand{\test}{\dataset_{\text{test}}}
\newcommand{\features}{\mathcal{X}}
\newcommand{\labels}{\mathcal{Y}}
\newcommand{\hypspace}{\mathcal{H}}
\newcommand{\params}{\bm{\theta}}
\newcommand{\Params}{\bm{\Theta}}
\newcommand{\w}{\bm{\omega}}
\newcommand{\loss}{\ell}
\newcommand{\Loss}{\mathcal{L}}
\newcommand{\ind}{\mathbbm{1}}

\newcommand{\gio}[1]{\todo[inline,color=teal!60]{{\bf Gio:} #1}}
\newcommand{\gabri}[1]{\todo[inline,color=red!60]{{\bf Gabri:} #1}}
\newcommand{\edo}[1]{\todo[inline,color=orange!60]{{\bf Edo:} #1}}

\section{Introduction}
\label{sec:intro}
The last decade has witnessed incredible advances in machine learning (ML) and artificial intelligence (AI), which in turn have led to the pervasive application of such techniques across several domains.
For example, ML/AI solutions are nowadays successfully used to solve complex tasks in natural language processing (NLP), computer vision, finance, and healthcare, to name a few.
One of the most remarkable successes of ML/AI is undoubtedly represented by recent \textit{multimodal} large language models (LLMs), such as GPT-4o~\cite{openai2024gpt4o_systemcard} and Gemini 1.5~\cite{geminiteam2024gemini15unlockingmultimodal}, as well as generative systems for rich media, including text-to-image models such as DALL-E 3~\cite{openai2023dalle3systemcard} and large-scale video generation models such as Sora~\cite{openai2024sora_systemcard}.
These tools have raised the bar of human-to-machine interaction to a \textit{new} level, unforeseeable only a few years ago.

The increasing complexity of deep neural network architectures at the heart of such tools, having millions or even billions of learnable parameters, requires a massive volume of data to avoid the risk of overfitting.
The standard approach for training such huge models is to collect large datasets in a single location, either physically centralized like a dedicated server or logically centralized like a cloud-based cluster of machines. 
Very often, this necessitates moving data from their local origins to a remote destination, causing at least two problems. 
First, the high-rate transmission of a large bulk of data may exhaust network bandwidth consumption, ultimately affecting energy efficiency, especially for resource-constrained devices like mobile smartphones.
Second, and most importantly, data ownership is transferred to a third-party entity with possible implications for its privacy and security.
This paradigm poses significant challenges in some critical application domains where ML/AI systems are deployed due to existing regulatory restrictions. Indeed, several initiatives, such as GDPR~\cite{gdpr-2018} and HIPAA~\cite{hipaa-1996}, have emerged to govern and protect the sharing of sensitive data. 
As it turns out, any company that (marginally) relies on ML/AI tools to run its business must satisfy data privacy and security requirements promulgated by current laws and regulations. 
This is even more stringent for so-called ``data-driven'' companies, whose revenue heavily depends on the value they are able to extract from data. 

Many solutions have been proposed in the literature to mitigate potential data privacy and security issues in the traditional ML setting, primarily leveraging \textit{differential privacy} (DP)~\cite{Dwork2014TheAF}, \textit{homomorphic cryptography} (HC)~\cite{8241854}, and \textit{secure multi-party computation} (SMPC)~\cite{smc,zhou2024smpc4ml} techniques. 
However, deploying these solutions in practice may be problematic. 
For instance, the amount of random noise injected by DP mechanisms must be carefully calibrated; adding more noise can guarantee higher data privacy protection, but might also decrease the model accuracy significantly.
Similarly, using HC to compute encrypted learning data may be limited to simple linear models~\cite{NWIJBT13} or a very small number of entities involved~\cite{6410315}. 
Finally, although SMPC can be used for large-scale ML, this technique is not immune to information leakage. 
Thus, instead of ``patching'' the standard ML framework with privacy-preserving solutions \textit{ex post}, the focus of the research community has shifted to planning novel ML training procedures that protect data privacy and security \textit{by design}.

As a result of such effort, \textit{federated learning} (FL) is a new ML paradigm first introduced by Google in 2017~\cite{google-fl-mcmahan-2017a} that addresses the challenges above. 
The authors advocate the need for a new \textit{distributed} training procedure with a well-known app for smartphones, i.e., the smart Google keyboard (Gboard). Indeed, at the core of Gboard, there is a neural language model that can predict, hence suggest, the next word to enter while the user is typing.
For such a model to be successful, it must be trained on vast text corpora collected from many user devices (e.g., smartphones). 
In contrast to the standard ML approach, which would require $(i)$ transferring these user-generated data from remote devices into Google's infrastructure, $(ii)$ training a model on those data at a central location, and $(iii)$ distributing the learned model back to user devices, Google proposed FL.
Generally speaking, FL trains a global model using distributed data, {\em without} the need for the data to be shared or transferred to any central facility.
In other words, FL takes advantage of a multitude of AI-enabled edge devices that cooperate with each other to \textit{jointly} learn a predictive model using their own local private data. 
Specifically, an FL system consists of a central server and many edge clients; a typical FL round involves the following steps: {\em(i)} the server randomly picks some clients and sends them the current, global model; {\em(ii)} each selected client locally trains its model with its own private data; then, it sends the resulting local model to the server;\footnote{Whenever we refer to global/local model, we mean global/local model {\em parameters}.} {\em(iii)} The server updates the global model by computing an \emph{aggregation function} on the local models received from clients (by default, the average, FedAvg~\cite{mcmahan2017googleai}). 
The process above continues until the global model converges.
Since this paradigm smoothly integrates with ubiquitous, distributed infrastructures, FL has been successfully applied to several domains, such as IoT~\cite{Wu20personalized}, Fog computing~\cite{Zhou20privacy}, autonomous vehicles~\cite{Pokhrel20decentralized}, and wearable devices~\cite{Chen20fedhealth}.

Although FL has unquestionable advantages over standard ML, it also presents problems related to centralized orchestration:
\begin{itemize}
    \item \textit{\textbf{Single Point of Failure:}} A centralized server may crash, causing the system to collapse or stall \cite{9464278}. 
    \item \textit{\textbf{Privacy, Trust Issues, and Security:}} In standard FL, the central server has access to the whole set of local models sent by clients, which may expose sensitive information through Model Inversion \cite{Fredrikson2015mia}, Membership Inference \cite{Hu2022membinfatt}, and Data Reconstruction~\cite{boenisch2023dratrapweights, zhu2019deepleakagegradients, zhao2020improveddeepleakage} attacks. Furthermore, the server could not only act as an honest but curious entity but could also be malicious as a whole, performing model poisoning attacks~\cite{baruch2019neurips, fang2020usenix, shejwalkar2021dnc} and data poisoning ones~\cite{Bagdasaryan2020backdoorfl, wang2020oodbackdoors, Xie2020DBA}. 
    \item \textit{\textbf{Communication Overhead and Latency:}} The server must communicate with an arguably growing number of clients, which may reach millions of devices in some cases~\cite{wang2021fieldguidefederatedoptimization}, leading to high bandwidth usage and latency~\cite{10.1145/3709021.3737667,zhu2021delayed}. 
    \item \textit{\textbf{Scalability Bottlenecks:}} As the number of clients increases, the central server may be unable to satisfy the growing system's needs due to its limited computational resources. 

\end{itemize}

We provide a comprehensive review of works that address \textit{decentralization of the orchestration/aggregation role} in FL, i.e., \textbf{\textit{decentralized} FL} (DFL). Accordingly, we exclude papers that primarily improve \emph{server-based} FL (e.g., compression, personalization, or secure aggregation under a coordinator) unless they are explicitly instantiated in a peer-to-peer or blockchain-based decentralized architecture. In DFL, the central coordinator is dropped, embracing a peer-to-peer architecture where nodes both train their own local model and aggregate the arriving ones from their neighbors. This solution eliminates the dependency on a single entity, providing advantages on all the four axes above: removing the single point of failure enhances the reliability of the system, and jeopardizing the entire network requires more effort on the adversarial side. Moreover, the computation and communication costs are spread across the entire network instead of a single server.

However, the benefits of decentralized FL do not come for free. First, peer-to-peer architectures must take into account the heterogeneity of their participants. Second, choosing the right network topology is essential to keep the bandwidth utilization low (e.g., a fully connected network is not more scalable than standard FL). Third, since there is no central server, peers are responsible of their own security and privacy. Fourth, incentivizing active participation from peers is essential to prevent lazy nodes. Overall, eliminating the single point of failure may inadvertently exacerbate some of the existing issues in standard FL systems.


\subsection{Comparison with previous surveys and contributions}
Previous surveys on DFL are largely axis-specific (Table~\ref{tab:survey-comparison}): \cite{dfl_survey_beltran_2023} emphasize fundamentals (FL architectures/topologies/communication), tooling and frameworks, KPIs, and application scenarios; \cite{dfl_sec_survey_Hallaji_2024} focus on security and privacy threats/defenses in DFL, with taxonomy and discussion primarily organized around the security–privacy lens; and \cite{dfl_survey_Liangqi_2024} propose multiple taxonomies centered on system design dimensions such as iteration order, communication protocol, network topology, and paradigm/temporal variability, aiming at IoT applications. In parallel, \cite{dfl_blockc_survey_ZHANG_2024} treats blockchain-based DFL as the primary object of study and organizes the field around ledger-centric concepts, frameworks, and challenges rather than comparing blockchain against non-ledger peer-to-peer decentralization on equal footing. All the aforementioned surveys, beyond being updated to cover the literature up to 2023, do not follow a systematic literature review methodology; as a result, their corpora may be affected by selection and reporting biases (e.g., over-representing highly visible keywords), which can skew the perceived maturity of the field and the conclusions drawn about open gaps.

In contrast, our survey contributes a unified, challenge-driven taxonomy that jointly covers traditional peer-to-peer DFL and blockchain-based DFL works, spanning the literature from the inception of DFL (2018) through 2026, and organizing both families by the core systems challenges they are designed to address (fault tolerance, privacy, security, incentives, heterogeneity, bandwidth efficiency, theory/task extensions, and resource efficiency). This perspective \textit{(i)} lets readers immediately retrieve “what to read” for a specific bottleneck, \textit{(ii)} exposes systematic coverage gaps (dimensions that are repeatedly under-addressed), and \textit{(iii)} makes the traditional vs. blockchain trade-space explicit within a single coherent map, rather than splitting it across architecture-only or security-only narratives.


In summary, the contributions of this article are as follows:
\begin{itemize}
    \item[\textit{(i)}] We discuss the fundamentals of standard FL, peer-to-peer systems, and blockchain, covering key aspects such as federation scale, data partitioning methods, communication architectures, challenges, and definitions.
    \item[\textit{(ii)}] We conduct a systematic literature review, selecting relevant works based on the \textit{Preferred Reporting Items for Systematic Reviews and Meta-Analyses} (PRISMA) guidelines.
    \item[\textit{(iii)}] We propose two novel taxonomies for standard peer-to-peer and blockchain-based DFL methodologies, categorizing and analyzing the most influential works based on the challenges they address.
    \item[\textit{(iv)}] We discuss open challenges and future directions for DFL systems.
\end{itemize}

\begin{table*}[!htb]
\centering
\caption{Comparison between our survey and prior surveys on DFL.}
\label{tab:survey-comparison}
\begin{adjustbox}{width=\textwidth}
\begin{tabular}{l c c c c c c c c c c}
\toprule
\textbf{Reference} &
\textbf{Year} &
\textbf{Lit. Coverage} &
\textbf{Scope / lens} &
\textbf{PRISMA} &
\textbf{Lit. Evolution} &
\textbf{Fundamentals} &
\textbf{Challenges Classification} &
\textbf{Pure P2P} &
\textbf{Blockchain} &
\textbf{Unified Taxonomy}
\\
\midrule

Martínez Beltrán \textit{et al.}~\cite{dfl_survey_beltran_2023} 
& 2023 & 2023
& Frameworks, KPIs, trends
& \Circle & \Circle & \LEFTcircle & \Circle & \CIRCLE & \Circle & \Circle \\

Hallaji \textit{et al.}~\cite{dfl_sec_survey_Hallaji_2024}
& 2024 & 2022
& Security \& privacy
& \Circle & \Circle & \LEFTcircle & \Circle & \LEFTcircle & \LEFTcircle & \Circle \\

Yuan \textit{et al.}~\cite{dfl_survey_Liangqi_2024}
& 2024 & 2023
& IoT-oriented, topologies
& \Circle & \LEFTcircle & \LEFTcircle & \Circle & \CIRCLE & \LEFTcircle & \Circle \\

Zhang \textit{et al.}~\cite{dfl_blockc_survey_ZHANG_2024}
& 2024 & 2023
& Blockchain-based DFL
& \Circle & \Circle & \LEFTcircle & \Circle & \Circle & \CIRCLE & \Circle \\

\midrule
\textbf{This survey}
& \textbf{2026} & \textbf{2026}
& \textbf{Comprehensive DFL landscape}
& \textbf{\CIRCLE} & \textbf{\CIRCLE} & \textbf{\CIRCLE} & \textbf{\CIRCLE} & \textbf{\CIRCLE} & \textbf{\CIRCLE} & \textbf{\CIRCLE} \\

\bottomrule
\end{tabular}
\end{adjustbox}

\vspace{0.4em}
\footnotesize \textbf{Legend:} \CIRCLE~fully covered; \LEFTcircle~partially covered; \Circle~not covered.
\end{table*}

The remainder of this paper is organized as follows. In Section~\ref{sec:background}, we recall some background and preliminary concepts.
Section~\ref{sec:method} describes the methodology used to collect the set of most relevant works on DFL proposed in the literature and considered in this survey. 
Section~\ref{sec:challenges} discusses the main challenges of DFL. 
We categorize and review the contributions in Section~\ref{sec:decentralized-fl}.
We also sketch several interesting lines of future research on this subject in Section~\ref{sec:future}.
Finally, Section~\ref{sec:conclusion} concludes our work.

\section{Background and Preliminaries}
\label{sec:background}
In this section, we review some background and preliminary concepts utilized throughout this manuscript. Specifically, we cover three main topics: $(i)$ \textit{federated learning} (FL), $(ii)$ \textit{peer-to-peer systems} (P2P), and $(iii)$ \textit{blockchain}. 

\subsection{Federated Learning}
\label{subsec:fl}
The prototypical FL setting consists of a central server $S$ and a set of distributed clients $\mathcal{C}$, such that $|\mathcal{C}|=K$, that jointly cooperate to solve a standard supervised learning task.\footnote{Notice that the FL paradigm can also be used to solve unsupervised learning tasks like K-means clustering~\cite{9499980}.} Each client $c\in \mathcal{C}$ has access to its own private training set $\dataset_c$, namely the set of its $n_c$ local labeled examples, i.e., $\dataset_c = \{\bm{x}_{c,i}, y_{c,i}\}_{i=1}^{n_c}$.

The goal of FL is to train a global predictive model whose architecture and parameters $\params^*\in \R^d$ are shared amongst all the clients and found to solve the following objective:
\begin{equation}
\label{eq:erm}
\params^* = \text{argmin}_{\params} \Loss(\params) = \text{argmin}_{\params} \sum_{c=1}^K p_c \Loss_c(\params;\dataset_c),
\end{equation}
where $\Loss_c$ is the local objective function for client $c$. Usually, this is defined as the empirical risk calculated over the training set $\dataset_c$ sampled from the client's local data distribution:
\begin{equation}
\label{eq:client-erm}
\Loss_c(\params;\dataset_c) = \frac{1}{n_c}\sum_{i=1}^{n_c} \loss(\params;(\insta_{c,i}, y_{c,i})),
\end{equation}
where $\loss$ is an instance-level loss (e.g., cross-entropy loss or squared error in the case of classification or regression tasks, respectively). 
Furthermore, each $p_c \geq 0$ specifies the relative contribution of each client. 
Since it must hold that $\sum_{c=1}^{K}p_c = 1$, two possible settings for it are: $p_c = 1/K$ or $p_c = n_c/n$, where $n = \sum_{c=1}^K n_c$. 

\begin{figure}[!htb]
\centering
\subfloat[A general FL round. (1) The server samples a subset of available clients, initializes the model, and sends it to the selected workers. (2) Receiving clients train the model on their local datasets (3) and send back the updated models to the server. (4) The server aggregates the local models using a specified aggregation rule.\label{fig:fl-arch}]{
  \includegraphics[width=0.95\linewidth]{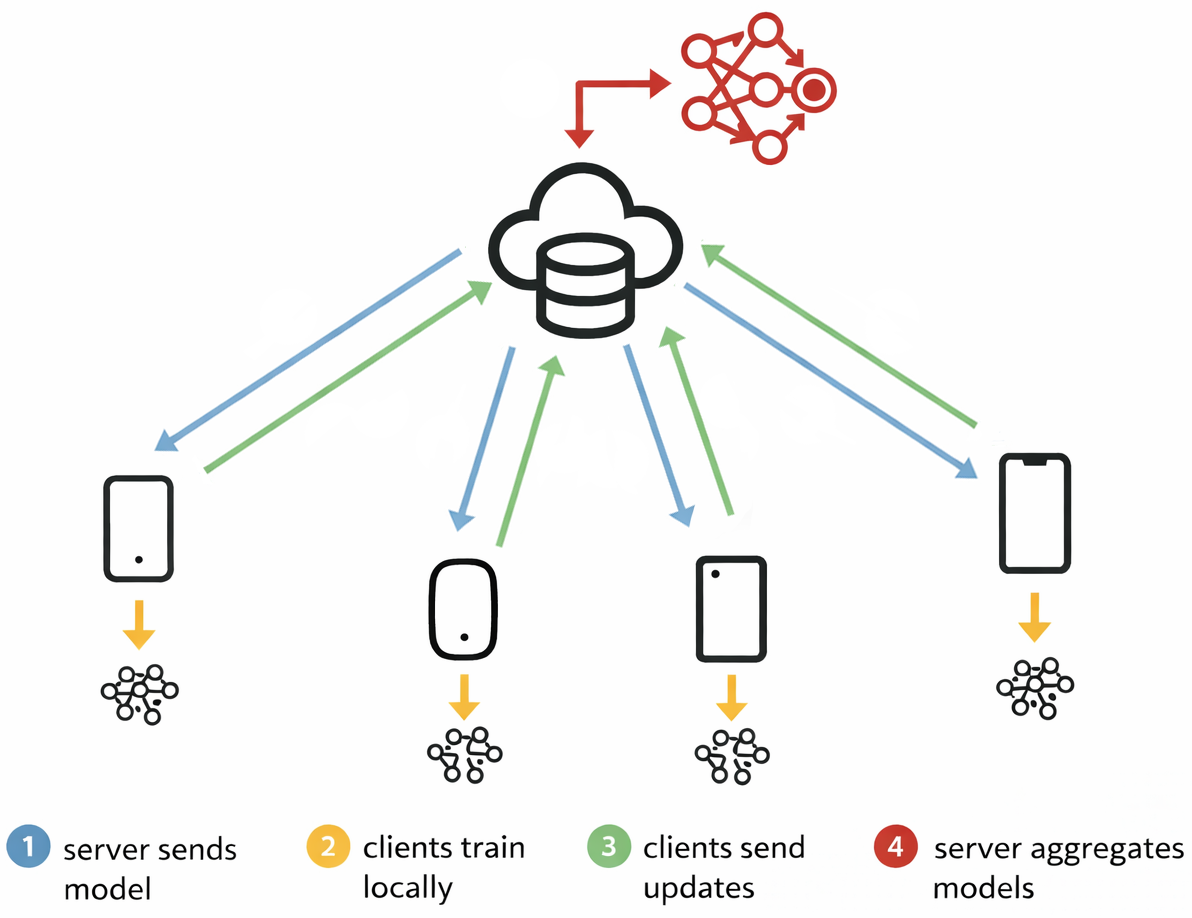}
}\\[0.8em]

\subfloat[A general decentralized FL (DFL) round. (1) Each node performs local training on its private data, (2) exchanges model updates with neighboring peers, and (3) updates its local model via neighbor-based local aggregation.\label{fig:dfl-arch}]{
  \includegraphics[width=0.95\linewidth]{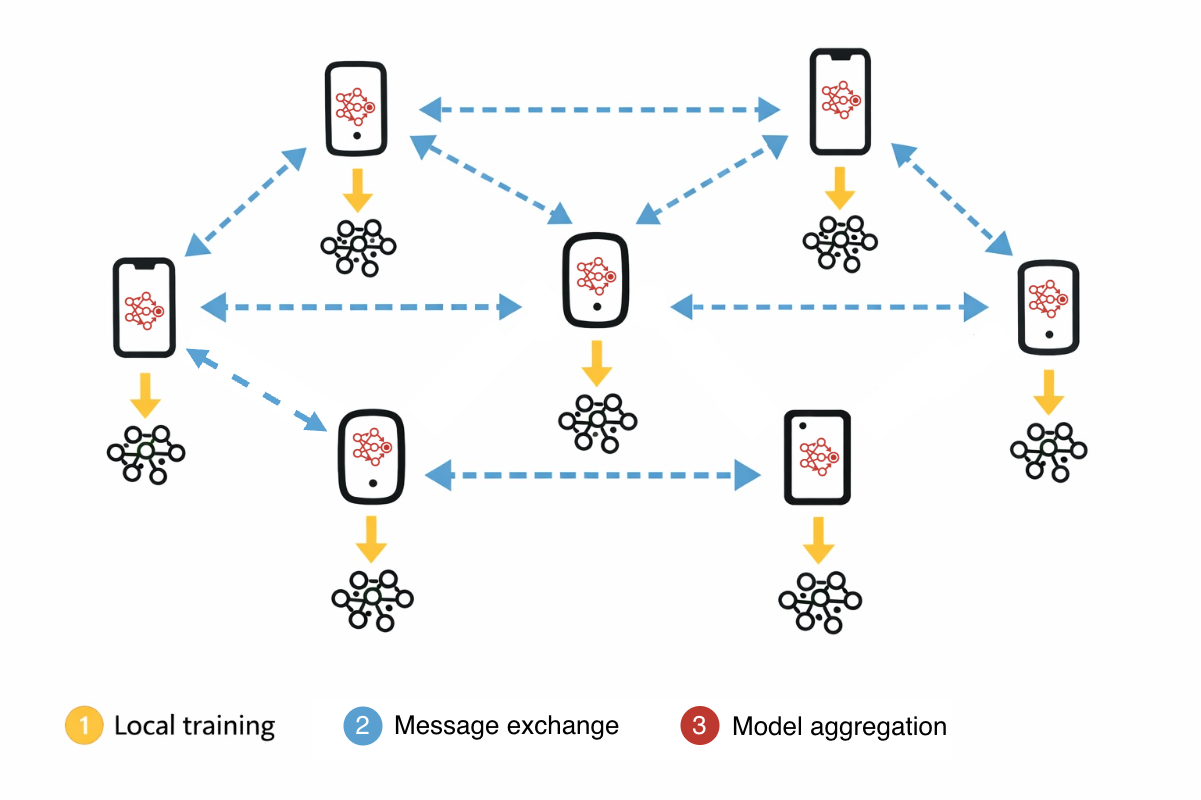}
}

\caption{Architectural comparison between standard FL and DFL. (a) FL relies on a central orchestrator for client selection and global aggregation. (b) DFL removes the central server and performs P2P coordination, where nodes iteratively train locally, exchange updates with neighbors, and aggregate received updates.}
\label{fig:architectural-comparison}
\end{figure}

The generic federated round at each time $t$ is decomposed into the following steps and iteratively repeated until convergence, i.e., for each $t=1,2,\ldots,T$:
\begin{enumerate}
\item[$(i)$] $S$ randomly selects a subset of clients ${\mathcal{C}}^{(t)}\subseteq \mathcal{C}$, so that $1 \leq |{\mathcal{C}}^{(t)}| \leq K$, and sends them the current, global model $\params^{(t)}$.
To ease of presentation, without loss of generality, in the following, we assume that the number of clients picked at each round is constant and fixed, i.e., $|\mathcal{C}^{(t)}| = m,~\forall t\in \{1,2,\ldots, T\}$.
\item[$(ii)$] Each selected client $c\in {\mathcal{C}}^{(t)}$ trains its local model $\params_c^{(t)}$ on its own private data $\dataset_c$ by optimizing the following objective, starting from $\params^{(t)}$:
\begin{equation}
    \params_c^{(t)} = \text{argmin}_{\params^{(t)}}\Loss_c(\params^{(t)}; \dataset_c).
\label{eq:client-opt}
\end{equation}
The value $\params_c^{(t)}$ is computed via gradient-based methods like stochastic gradient descent (SGD) and sent back to $S$.
\item[$(iii)$] $S$ computes $\params^{(t+1)} = \phi(\{\params_c^{(t)}~|~c\in \mathcal{C}^{(t)}\})$ as the updated global model, where $\phi: \R^{d^m} \mapsto \R^d$ is an \emph{aggregation function}; for example, $\phi = \frac{1}{m}\sum_{c\in \mathcal{C}^{(t)}} \params_c^{(t)}$, i.e., FedAvg or one of its variants~\cite{lu2020spml}.
\end{enumerate}
Figure~\ref{fig:fl-arch} illustrates this standard client–server FL workflow.

A few alternatives to the scheme above are possible. For example, in step $(ii)$, instead of sending the vector of parameters $\params_c^{(t)}$, each selected client could transmit to the server its displacement vector compared to the global model received at the beginning of the round, i.e., $\bm{u}_c^{(t)} = \params_c^{(t)} - \params^{(t)}$. 
This way, in step $(iii)$, the server will compute the new global model as $\params^{(t+1)} = \params^{(t)} + \phi(\{\bm{u}_c^{(t)}~|~c\in \mathcal{C}^{(t)}\})$, namely the aggregation function is calculated on the local update vectors rather than the actual local models.
Similarly, sending local models (or their updates) $\params_c^{(t)}$ ($\bm{u}_c^{(t)}$) is equivalent to sending ``raw'' gradients $\nabla \Loss_c^{(t)}$ to the central server; in the latter case, $S$ simply aggregates the gradients and uses them to update the global model, i.e., $\params^{(t+1)} = \params^{(t)} - \eta \phi(\{\nabla \Loss_c^{(t)}~|~c\in \mathcal{C}^{(t)}\})$, where $\eta$ is the learning rate.

Broadly speaking, FL systems can be categorized according to four different axes: \textit{data partitioning}, \textit{machine learning model}, \textit{scale of federation}, and \textit{communication architecture}. Below, we review each of these aspects separately.

\subsubsection{Data Partitioning}

Data can be distributed over the sample and feature spaces, which creates a categorization in \textit{horizontal}, \textit{vertical}, and \textit{hybrid} FL~\cite{10.1145/3298981}. 

In \textit{horizontal} FL, the datasets of different clients have the same feature space but little intersection on the sample space. The majority of studies on FL assume horizontal partitioning. Since the local data are in the same feature space, the parties can train the local models using their local data with the same model architecture, and the global model can simply be updated by averaging all the local models using standard FedAvg. 
A typical example where horizontal FL comes into play is when two (or more) regional branches of a bank want to collaboratively train a loan prediction model for their customers. Each branch may have very different user groups due to its geographical location, and the common set of users shared between any two branches is very small. However, their business is very similar, so the feature spaces are likely the same.

In \textit{vertical} FL, instead, the datasets of different clients have the same or similar sample space but differ in the feature space. It usually adopts entity alignment techniques~\cite{10.5555/2344108} to collect the overlapped samples of the parties, and then these data are used to train the machine learning model using encryption methods. 
As an example, consider two different companies in the same city, e.g., a bank and an e-commerce business. Their customer bases are likely to contain most of the residents of the area, so the intersection of their user space is large. However, the bank may record the user's revenue and expenditure behavior and credit rating, whereas the e-commerce collects the user's browsing and purchasing history; thus, their feature spaces are different.

Finally, \textit{hybrid} FL (also referred to as \textit{federated transfer learning} or FTL) is a combination of horizontal and vertical data partitioning. FTL applies when any pair of datasets from federated clients differ both in samples and in feature space. For instance, consider two different companies that are also geographically distant from each other, e.g., a bank located in China and an e-commerce company located in the United States.
Due to geopolitical restrictions, the user groups of the two companies have a small intersection.
Moreover, due to the different businesses, only a small portion of the feature space from both parties overlaps. In this case, \textit{transfer learning}~\cite{pan2010tkde} techniques can be applied to provide solutions for the entire sample and feature space under a federation.

\subsubsection{Machine Learning Models}
FL is primarily used to jointly solve a machine learning task, which usually consists of collaboratively training a common model on several distributed private datasets.
The choice of the specific ML model to train depends on the problem at hand and the dataset. 
The most popular family of models used within FL systems is Neural Networks (NNs) in all their flavors. Simpler models, such as Linear Regression (LinReg) or Logistic Regression~\cite{7958569} (LogReg), and Decision Trees~\cite{10632132} (DTs) or ensembles of those like the Gradient Boosting Decision Trees (GBDTs) are also successfully proposed in FL environments~\cite{DBLP:journals/corr/abs-1711-10677, Li_Wen_He_2020}, mostly due to their high efficiency and interpretability.

Generally speaking, an FL system can consist of homogeneous or heterogeneous ML models. In the former case, all clients have the same model, and aggregation of gradients comes into play at the server. In the latter scenario, there is no need for aggregation since each client has a different model. Therefore, at the server's end, aggregation methods are replaced by ensemble methods~\cite{yuFederatedLearningAlgorithm2023} like majority voting~\cite{10154565}, or knowledge distillation techniques~\cite{10682872}.

\subsubsection{Scale of Federation}
FL systems can be categorized into \textit{cross-silo} and \textit{cross-device}~\cite{9464278}. 
The difference between the two concerns the number of parties involved and the amount of data available at each client.
The simplest way to understand them is to associate cross-silo with large organizations or data centers and cross-devices with mobile devices. 
In the case of cross-silo, the number of federated clients is usually small, but they have extensive computational abilities (e.g., a group of large medical institutions). 
The main challenge of cross-silo FL is to create an efficient distributed computation under the constraint of privacy~\cite{DBLP:journals/tpds/ZhouXGHZM19}. 
When it comes to cross-device, instead, FL systems are made of a massive number of clients, each one with limited computational power (e.g., Google Gboard). 
In cross-device FL, the primary challenge relates to devices' energy consumption, which limits the complexity of training tasks that they can perform. 

\subsubsection{Communication Architecture}
The centralized architecture follows a client-server model, where a central entity acts as the \textit{orchestrator}. Its role involves coordinating the entire distributed training process, including aggregating individual local models sent by clients into a single global model.
On the other hand, in a decentralized architecture, there is no fixed distinction between client and server roles. Instead, each participant can both train on local data and aggregate models received from its neighbors; in some designs, aggregation responsibilities may rotate or be assigned to a subset of nodes. We contrast the client–server FL pipeline with its decentralized counterpart in Figure~\ref{fig:dfl-arch}.

Implementing \textit{decentralized} FL systems poses significant challenges and can be categorized into two main categories: traditional distributed computing methods, such as P2P approaches, and those utilizing blockchain technology.\\
In this survey, we will examine and assess the most relevant approaches to decentralized FL that have been proposed in the existing literature.

\subsection{Peer-to-Peer Systems (P2P)}
\label{subsec:p2p}
Generally speaking, P2P computing offers an alternative to the traditional client-server architecture~\cite{barkai2000introduction}. In a client-server model, clients connect to a central server to make requests while the server processes and responds to those requests. In contrast, P2P distributed systems allow nodes in the network to act as both servers and clients, promoting decentralization.

Although it is customary to associate P2P systems with (illegal) content sharing, they have broader applications beyond that. Increasingly, P2P solutions are being deployed to address various problems that were traditionally reliant on centralized server-based approaches. Some examples of P2P systems are BitTorrent~\cite{torrent}, Gnutella~\cite{gnutella}, Napster~\cite{napster}, and Skype~\cite{skype}.
The fundamental principle of P2P networks is to enable resource sharing among end systems, including files, storage space, CPU cycles, and more. These systems create an overlay network, facilitating communication between peers.

There are three primary characteristics of a P2P system~\cite{Roussopoulos20032PO}:
\begin{itemize}
    \item \textit{\textbf{Self-organizing}}: Nodes must organize themselves to form an overlay network. There should be no assistance from a central node. Also, there should not be any global index that lists all the peers and/or the available resources.
    \item \textit{\textbf{Symmetric communication}}: All nodes must be equal (i.e., no node should be more important than any other node). Also, peers should both request and offer services (i.e., they should act as both clients and servers).
    \item \textit{\textbf{Decentralized control}}: There should not be a central controlling authority that dictates behavior to individual nodes. Peers should be autonomous and must determine their level of participation in the network on their own.
\end{itemize}

P2P networks offer some advantages over classical client-server architectures, such as eliminating the single-point-of-failure and single-source bottleneck. 
The primary characteristics of P2P networks are \textit{reliability}~\cite{LI201020} against nodes that disconnect or have a low latency or bandwidth; \textit{scalability}~\cite{article1} as the workload is no more concentrated in a server; \textit{privacy}~\cite{10.1145/1851182.1851198} and \textit{anonimity}~\cite{1231513} via cryptographic protocols. 

Still following~\cite{Roussopoulos20032PO}, some of the key challenges of P2P systems are as follows:
 \begin{itemize}
     \item \textit{\textbf{Budget:}} The budget allocated to a solution influences the consideration of P2P architectures. If the budget is ample, the inefficiencies and complexities associated with P2P may not be deemed worthwhile. However, if the budget is limited, the low cost of entry for individual peers becomes an attractive factor despite the increased total system cost. Utilizing local components and surplus resources may be a justifiable approach within constrained budgets.
    \item \textit{\textbf{Resource relevance to participants:}} The relevance of data to peers plays a significant role in P2P cooperation. If the probability of peers being interested in each other's data is high, cooperation naturally evolves. Conversely, if relevance is low, artificial or extrinsic incentives may be necessary to foster cooperation.
    \item \textit{\textbf{Trust:}} Trust among peers varies depending on the specific problem requirements. Mutual distrust can be either essential or negligible. However, the cost of mutual distrust in P2P systems is high, and its necessity must be justified based on the problem's characteristics.
    \item \textit{\textbf{Rate of system change:}} P2P systems may experience stable or rapidly changing participants, resources, and parameters. Rapid changes pose challenges in ensuring consistency guarantees, defending against flooding, and mitigating other attacks.
    \item \textit{\textbf{Criticality:}} If the problem being solved is critical to users, centralized control may be demanded regardless of technical criteria. Even when P2P is not ruled out, the need for expensive security measures or extensive over-provisioning may render it economically unfeasible.
    \item \textit{\textbf{Security:}} P2P networks are known to be vulnerable to various types of security attacks. For instance, malicious nodes can disrupt the network by flooding it with redundant data, manipulating trust values in trust-based systems, coordinating with other malicious nodes for distributed denial-of-service (DDoS) attacks, or performing \textit{Sybil} attacks by creating multiple fake identities within the network~\cite{pretre2005attacks}.
 \end{itemize}

In recent years, \textit{decentralized} machine learning has gained popularity within P2P networks. The focus has been on solving the distributed consensus problem, aiming to find a global model that minimizes the sum of local loss functions~\cite{article3, article2}. Additionally, research has explored privacy-preserving approaches and scenarios where agents have distinct objectives~\cite{pmlr-v84-bellet18a}. These advancements enable collaborative learning and optimization in a decentralized manner.

\subsection{Blockchain}
\label{subsec:Blockchain}
In essence, a \textit{blockchain}  is an open and distributed ledger designed to record transactions between parties securely and permanently. It operates by facilitating the transfer of digital assets (e.g., cryptocurrencies) from one account to another. Transactions are packed into \textit{blocks}, forming a chain where each block is linked to its predecessor using hash functions. Consensus mechanisms are used to achieve agreement within the blockchain network. According to~\cite{Yaga2018BlockchainTO}, some key properties of this technology are:
\begin{itemize}
\item \textit{\textbf{Ledger}}: Append-only ledger to provide full transactional history. Unlike traditional databases, transactions and values in a blockchain are not overridden.
\item \textit{\textbf{Secure}}: Cryptographically secure, the data within the ledger is attestable.
\item \textit{\textbf{Shared}}: The ledger is shared amongst multiple participants. This provides transparency across the node participants in the blockchain network.
\item \textit{\textbf{Distributed}}: The blockchain  can be distributed. By increasing the number of nodes, the ability of a bad actor to impact the consensus protocol used by the blockchain is reduced.
\end{itemize}

Blockchain technology has had a recent impact, although it dates back 30 years. In 2008, a paper published pseudonymously by Satoshi Nakamoto~\cite{nakamoto2009bitcoin} described \textit{Bitcoin}: a P2P electronic cash system that works like a chain of digital signatures.
Each owner transfers the coin to the next by digitally signing a hash of the previous transaction and the public key of the next owner and adding these to the end of the coin. 
A payee can verify the signatures to verify the chain of ownership. 
The consensus algorithm to decide which is the next block is called \textit{proof-of-work} (PoW), which is, in brief, a computationally expensive puzzle.
Once a node solves the puzzle, it packages the transactions into a block, attaches the PoW solution, and broadcasts the full block to the network. Other nodes adopt the longest chain rule to maintain consensus.

After the introduction of Bitcoin, the concept of blockchain expanded to include programmable capabilities. In 2015, Ethereum~\cite{Buterin2013} was introduced as a blockchain platform with a built-in programming language, allowing the creation of smart contracts and decentralized applications. 
Ethereum uses its cryptocurrency, called ``Ether'' as the internal fuel for executing transactions and running smart contracts. In Ethereum, there are two types of accounts: externally owned accounts controlled by private keys and contract accounts controlled by their contract code. Smart contracts enable users to define rules for ownership, transaction formats, and state transitions. The consensus protocol used in Ethereum is called \textit{proof-of-stake} (PoS), where validators stake their capital in Ethereum to propose and add new blocks to the blockchain. Validators are incentivized to follow the rules, as deviations can result in penalties or loss of stake.

Blockchain technology has also found utility in various ML tasks~\cite{8622598}. Researchers have explored different consensus algorithms that are particularly relevant to FL. These include \textit{proof-of-federated-learning} (PoFL), which repurposes the energy consumed in traditional PoW algorithms for solving meaningful puzzles in the context of FL \cite{9347812}. 
Moreover, \textit{proof-of-quality} (PoQ) introduces a consensus protocol based on Quality-of-Service (QoS) to ensure high-quality contributions to the blockchain~\cite{article4}. 
In addition, \textit{proof-of-authority} (PoA) relies on identity as a stake to achieve faster transaction processing~\cite{Liu2019MDPBasedQA}. 
Finally, committee-based consensus algorithms involve the validation of local gradients before appending them to the blockchain, where a committee of honest nodes verifies and generates blocks~\cite{9293091}. 

Despite these advantages, the literature consistently highlights that blockchain deployments introduce substantial overheads and non-trivial trade-offs, which can outweigh the benefits when the application does not strictly require a shared, tamper-evident ledger. The main challenges and downsides discussed in prior work can be summarized as follows:
\begin{itemize}
    \item \textit{\textbf{Scalability and throughput limits:}} Ledgers typically support limited transaction throughput; as workload increases, networks may experience congestion, longer confirmation times, and degraded performance~\cite{Aldoubaee2023, 10.1145/3700641}.
    \item \textit{\textbf{Latency and slow finality:}} Consensus and global replication increase end-to-end delay and make finality slower than in centralized databases~\cite{10.1145/3700641}.
    \item \textit{\textbf{Resource and energy overhead:}} Consensus can impose high computational and energy costs (especially under PoW), while other protocols reduce energy consumption at the expense of additional coordination assumptions and communication complexity~\cite{An_2023}.
    \item \textit{\textbf{Storage and communication overhead:}} Maintaining a replicated ledger entails persistent storage growth and sustained bandwidth usage for block dissemination and validation~\cite{11029213}.
    \item \textit{\textbf{Security risks beyond cryptography:}} While transactions are cryptographically protected, real systems remain vulnerable to consensus-level attacks (e.g., majority/51\% attacks in weakly provisioned networks), and smart-contract vulnerabilities, any of which can compromise integrity or availability~\cite{10.1145/3700641}.
    \item \textit{\textbf{Privacy tensions:}} Immutable and transparent ledgers can conflict with data-protection requirements (e.g., minimization and erasure rights) and increase the compliance burden in regulated domains~\cite{VillanuevaBlockchainTA}.
\end{itemize}
In DFL environments, where peers already perform the core functions of local training and neighbor-based aggregation, these overheads must be carefully weighed before introducing a blockchain layer. In other words, blockchain should not be treated as a default substrate for decentralization, but as an \emph{optional} component justified by concrete requirements (e.g., auditability, accountability, or incentive settlement) that cannot be met with lighter-weight peer-to-peer coordination and cryptographic primitives alone.
\section{Research Method}
\label{sec:method}
We have chosen to follow the \textit{Preferred Reporting Items for Systematic Reviews and Meta-Analyses} (PRISMA) guidelines~\cite{page2021prisma} for conducting our review. 
Our approach consists of three primary steps. 
\begin{enumerate}
    \item[$(i)$] \textit{\textbf{Identifying relevant works}}: We find relevant articles by conducting a comprehensive literature search and screening titles, abstracts, and full-text articles. 
    \item[$(ii)$] \textit{\textbf{Appraising study quality}}: We validate the quality of the included studies, taking into account factors such as study design, methodology, and potential biases. 
    \item[$(iii)$] \textit{\textbf{Synthesizing findings and drawing conclusions}}: We synthesize the findings from the included studies and draw meaningful conclusions. 
\end{enumerate}
By following these steps, we aim to ensure a rigorous and comprehensive review process.

\subsection{Research Questions}
In this survey, we begin by formulating a concise set of \textit{research questions} (\textbf{RQs}) to guide our review of the DFL literature. The RQs are reported below as a list and are answered throughout the paper to structure the discussion and ensure consistent coverage of the main aspects considered in our taxonomy and comparative analysis.

\begin{itemize}
    \item \textbf{Motivation and problem framing}
    \begin{itemize}
        \item \textbf{RQ1:} What is the centralized orchestration problem in FL, and what changes when the server is removed?
    \end{itemize}

    \item \textbf{Taxonomy and design space}
    \begin{itemize}
        \item \textbf{RQ2:} What are the main architectural families of decentralized FL, and how can existing approaches be organized under a unified taxonomy?
    \end{itemize}

    \item \textbf{Empirical practice}
    \begin{itemize}
        \item \textbf{RQ3:} How is DFL evaluated in the literature, and what gaps emerge from current experimental practice?
    \end{itemize}

    \item \textbf{Challenges, lessons, and directions}
    \begin{itemize}
        \item \textbf{RQ4:} What are the key challenge dimensions in decentralized FL, lessons and future directions?
    \end{itemize}
\end{itemize}
\noindent\underline{Where these RQs are addressed.}
\textbf{RQ1} is addressed in Sec.\ref{sec:intro} (motivation and limits of centralized orchestration) and revisited in Sec.\ref{sec:challenges} (challenge implications).
\textbf{RQ2} is addressed in Sec.\ref{sec:decentralized-fl} (two-class taxonomy TD-FL vs BC-FL and challenge-driven organization) and supported by Fig.~\ref{fig:timeline}, Table~\ref{tab:taxonomy}, and Fig.~\ref{fig:taxonomy}.
\textbf{RQ3} is addressed by Table~\ref{tab:experimental-settings} and the accompanying discussion on evaluation coverage and limitations.
\textbf{RQ4} is addressed in Sec.\ref{sec:challenges} (challenge dimensions), Sec.~\ref{sec:decentralized-fl} (mapping works to dimensions), and Sec.~\ref{sec:future} (open directions and gaps).

\subsection{Search Process}
For the purpose of this review, we conducted a search within a specific time frame spanning from 2018 to early 2026, with a focus on DFL. 
We utilized several common databases, including IEEE Xplore, ScienceDirect, ACM Digital Library, and Google Scholar, employing advanced search options and boolean expressions (AND, OR).

\begin{figure}[!htb]
    \centering
    \includegraphics[width=0.9\linewidth]{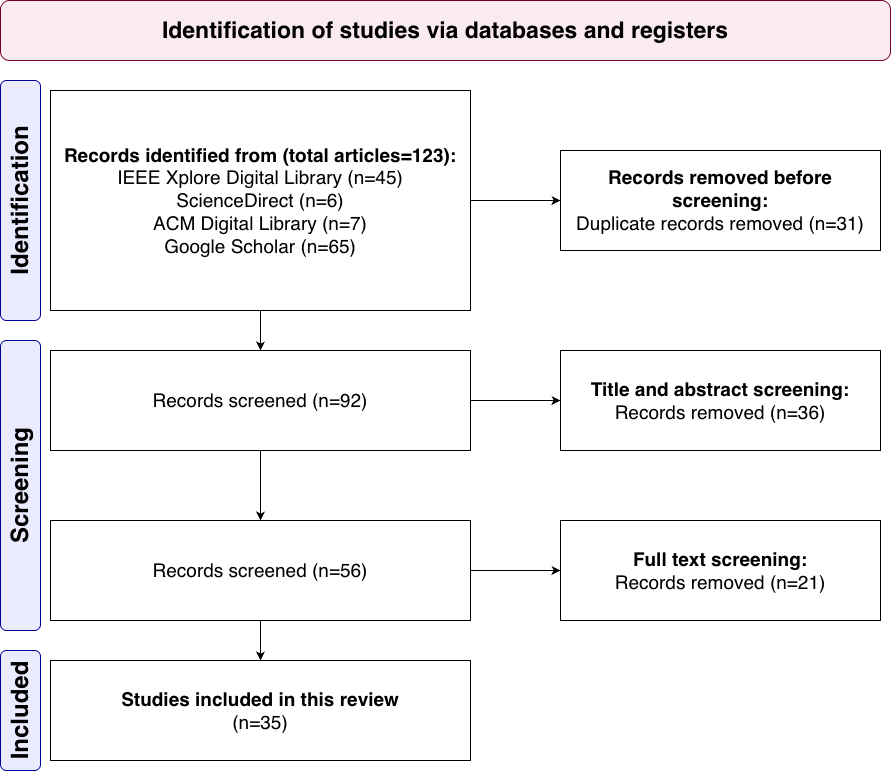}
    \caption{Selection process of articles with the PRISMA flow diagram.}
    \label{fig:prisma}
\end{figure}

We paid special attention to the exclusion of articles that did not align with the scope of this survey. 
To provide an overview of the article selection process and the statistics of articles considered in this survey, we present the PRISMA flow diagram in Fig.~\ref{fig:prisma}. Each stage of the diagram is discussed in detail in the following sections.

\subsection{Inclusion/Exclusion Criteria}
The criteria prioritize the inclusion of articles that $(i)$ address the centralized orchestration problem, $(ii)$ clearly define themselves as decentralized frameworks, $(iii)$ delegate aggregation to peers rather than a central entity, and \textit{(iv)} propose a substantive architectural or algorithmic novelty beyond incremental variants.

During the keyword search, articles were collected based on the presence of specific terms, even if mentioned only once in the paper. However, articles that did not meet the predetermined criteria or were deemed irrelevant were excluded from the review. Examples of excluded articles encompass those that lacked decentralization or exhibited hybrid characteristics, studies that did not focus on decentralization and centralized orchestration problems, or those that lacked novelty.
The inclusion and exclusion process is depicted in Fig.~\ref{fig:prisma}.

A total of 123 articles were retrieved from the specified databases. After removing duplicates, 92 articles remained for further analysis. These articles underwent two rounds of screening. Initially, the titles and abstracts were meticulously reviewed, resulting in the exclusion of 36 articles. Subsequently, the full texts of the remaining 56 articles were thoroughly examined, leading to the disqualification of an additional 21 papers. Ultimately, 35 articles were identified from the initial pool of 123 articles and included in the final review.

\subsection{Data Extraction}
The data collection process was centered on addressing the research questions of our study. Each of the 35 articles was individually examined, and distinct data points were extracted to inform our findings. The collected data were systematically recorded in a spreadsheet. The following information was extracted from each article:
\begin{enumerate}
    \item[$(i)$] Document title, publication year, and the name of the journal/conference in which it was published.
    \item[$(ii)$] Details of the datasets used and the federated settings in which they were employed.
    \item[$(iii)$] Identification of whether the implementation utilized ``standard'' distributed computing approaches or blockchain technology.
    \item[$(iv)$] The algorithms employed for training ML models.
    \item[$(v)$] Performance metrics, main results, and limitations.
\end{enumerate}
By gathering this information, we aimed to obtain a comprehensive understanding of the articles' content and to extract relevant data points for analysis to address our research questions effectively.
\section{Challenges of Decentralized FL}
\label{sec:challenges}

In this section, we discuss the primary challenges of DFL, which we categorize into the following groups: $(i)$ \textit{fault tolerance}, $(ii)$ \textit{bandwidth utilization}, $(iii)$ \textit{data heterogeneity}, $(iv)$ \textit{adversarial attacks}, and $(v)$ \textit{lack of incentive mechanisms}.

\subsection{Fault Tolerance}
By eliminating reliance on a central server, DFL solves the single point of failure problem. However, its decentralized nature introduces unique challenges in fault tolerance, i.e., the system’s ability to recover from client failures or network disruptions.

Participants operate in dynamic environments where device disconnections, power constraints, or network latency lead to partial participation in training rounds. Unlike centralized FL, where a server orchestrates updates, DFL’s P2P structure and intrinsic unreliability of nodes complicate recovery from such failures. For instance, in a network resembling a line topology, if a node located in the middle fails, the network splits until the node becomes available again. Consequently, neighboring participants may not receive model updates for several rounds~\cite{dfl_survey_Liangqi_2024}.

\subsection{Bandwidth Utilization}
Scalability in DFL hinges on efficient communication across participating nodes. In fully connected topologies, each client communicates with all peers, leading to $O(n^2)$ message complexity for $n$ clients (compared to $O(n)$ cost for a standard FL settings), a prohibitive cost for large networks.

\subsection{Data Heterogeneity}
In the context of FL, non-IID data refers to scenarios where the data distribution varies significantly across participating clients. Unlike traditional centralized ML, where training data typically follows a uniform distribution, FL systems operate in environments where each client generates or collects data based on unique user behaviors, preferences, and contexts. This inherent heterogeneity manifests in various forms of distribution skew. For instance, in cross-device scenarios involving smartphones or IoT devices, each unit captures data within its distinct operational environment. Similarly, in cross-silo applications involving different organizations or departments, the data often exhibits significant distribution differences based on organizational functions and user.

Recent studies have identified that inconsistencies in client loss landscapes primarily cause performance degradation in non-IID scenarios~\cite{seo2025understandingfederatedlearningiid}. When clients optimize their local models on divergent data distributions, they navigate different optimization trajectories, leading to conflicting update directions that hinder global convergence. This phenomenon, often termed \textit{client drift}, causes local models to become increasingly specialized to their particular data distributions while diverging from the optimal global solution.

The fundamental optimization objective in FL involves finding parameters that perform well across all client distributions simultaneously. However, with non-IID data, the optimization landscape becomes more complex, featuring multiple local minima that may be optimal for individual clients but suboptimal for the collective. This creates tension between local optimization and global performance, particularly in decentralized settings where no central coordinator exists to harmonize these conflicting objectives~\cite{seo2025understandingfederatedlearningiid}.

\subsection{Adversarial Attacks and Privacy}
Traditional FL faces a range of adversarial attacks. These attacks manifest in different forms and can adversely affect output accuracy, disrupt model convergence, and even result in unacceptable denial-of-service (DoS) for both servers and users. Similar threats are also shared by DFL approaches.

One major challenge is the presence of possibly \textit{malicious} nodes. These intentionally join the network to exploit P2P networking characteristics and harm the system. They may flood other nodes with random or redundant data and collaborate with other malicious actors to execute DoS attacks. These malicious nodes pose a significant security and reliability challenge to the DFL process; thus, if designers do not impose access control (or equivalently, if the access control protocol is faulty), any malicious actor can join the network and tamper with the learning process.

The creation of federated models involves aggregating model updates submitted by participants. However, to protect the confidentiality of training data, the aggregator has no visibility into how these updates are generated. This limitation makes DFL susceptible to several kind of poisoning attacks that aim to detriment the accuracy of the global model by either polluting the training set of some nodes with maliciously crafted examples (\emph{data poisoning} \cite{Bagdasaryan2020backdoorfl, wang2020oodbackdoors, Xie2020DBA, 8418594}), or by directly manipulating the model weights before they arrive to the aggregator (\emph{model poisoning} \cite{baruch2019neurips, fang2020usenix, shejwalkar2021dnc, pmlr-v97-bhagoji19a}). Furthermore, while in standard FL only the server has access to the model parameters sent by the clients, in DFL, every participant has access to them. This can be exploited by malicious actors to carry on Membership Inference~\cite{Hu2022membinfatt} or Data Reconstruction~\cite{boenisch2023dratrapweights, zhu2019deepleakagegradients, zhao2020improveddeepleakage} attacks, where the private training set can be inferred from model parameters, posing significant risks to the users' privacy.

\subsection{Incentive Mechanisms}
Another problem is the lack of an incentive mechanism for the devices that participate in the DFL task. In a large-scale FL system, all devices want to participate in the learning process and improve their local model with the help of a federation. Since the aggregation of local updates costs computation power, it is necessary to choose devices with high-quality data and high computation power. Overall, incentives in FL serve as catalysts for addressing resource constraints, security concerns, data quality, and participant engagement. They help create a mutually beneficial environment where participants are motivated to contribute their resources and data, leading to improved training performance and the overall success of FL. Furthermore, since we have cross-silo and cross-device systems as defined in Section \ref{sec:background}, we have different reasons to create an incentive mechanism.

There are two main reasons why incentives are useful in cross-device FL~\cite{yang2022practical}. Firstly, privacy concerns and reluctance to dedicate computing resources can discourage mobile users from participating in DFL. Secondly, the current DFL approach rewards participants with the same global model, regardless of their contributions. This lack of differentiation may discourage active participation. Incentives can address these issues by providing reassurance regarding privacy and compensating participants based on their level of contribution, thereby motivating more active and engaged involvement in DFL. 

In cross-silo FL~\cite{tang2021incentive}, organizations can choose their processing capacity for local training. This decision affects the accuracy of the global model and the computational costs incurred. Communication costs depend on the frequency of model updates exchanged with the central server. However, organizations may have different valuations of precision and varying computational and communication costs. To promote efficient cooperation and maximize social welfare, an incentive mechanism is needed to motivate organizations, even if they are independent and driven by self-interest. This mechanism should align individual interests to maximize overall welfare in cross-silo FL.

\section{A Taxonomy of decentralized FL Approaches}
\label{sec:decentralized-fl}
\begin{figure*}[!htb]
\centering
\begin{tikzpicture}[
    yearbox/.style={
        rectangle,
        rounded corners=3pt,
        minimum height=0.6cm,
        minimum width=1cm,
        draw=black,
        line width=0.7pt,
        fill=white,
        font=\bfseries\footnotesize,
        drop shadow={shadow xshift=0.6pt, shadow yshift=-0.6pt, opacity=0.3}
    },
    approach/.style={
        rectangle,
        rounded corners=2pt,
        minimum height=0.35cm,
        minimum width=0.75cm,
        draw=black!60,
        line width=0.4pt,
        fill=#1,
        font=\tiny\sffamily,
        text=black,
        drop shadow={shadow xshift=0.3pt, shadow yshift=-0.3pt, opacity=0.2},
        align=center,
        inner sep=1pt
    }
]
\draw[line width=3pt, blue!30, rounded corners=2pt] (0,0.78) -- (13,0.78);
\foreach \x in {0,1.625,3.25,4.875,6.5,8.125,9.75,11.375,13} {
    \fill[blue!60!black] (\x,0.78) circle (3pt);
}
\draw[line width=3pt, purple!30, rounded corners=2pt] (0,-0.78) -- (13,-0.78);
\foreach \x in {0,1.625,3.25,4.875,6.5,8.125,9.75,11.375,13} {
    \fill[purple!60!black] (\x,-0.78) circle (3pt);
}
\node[font=\bfseries\small\sffamily, text=blue!60!black] at (6.5,1) {Traditional Distributed FL};
\node[font=\bfseries\small\sffamily, text=purple!60!black] at (6.5,-1) {Blockchain-based FL};

\draw[line width=0.6pt, blue!50!black] (0,0.3) -- (0,0.78);
\draw[line width=0.6pt, purple!60!black] (0,-0.3) -- (0,-0.78);
\node[yearbox] at (0,0) {2018};

\draw[line width=0.6pt, blue!50!black] (1.625,0.3) -- (1.625,0.78);
\draw[line width=0.6pt, purple!60!black] (1.625,-0.3) -- (1.625,-0.78);
\node[yearbox] at (1.625,0) {2019};

\draw[line width=0.6pt, blue!50!black] (3.25,0.3) -- (3.25,0.78);
\draw[line width=0.6pt, purple!60!black] (3.25,-0.3) -- (3.25,-0.78);
\node[yearbox] at (3.25,0) {2020};

\draw[line width=0.6pt, blue!50!black] (4.875,0.3) -- (4.875,0.78);
\draw[line width=0.6pt, purple!60!black] (4.875,-0.3) -- (4.875,-0.78);
\node[yearbox] at (4.875,0) {2021};

\draw[line width=0.6pt, blue!50!black] (6.5,0.3) -- (6.5,0.78);
\draw[line width=0.6pt, purple!60!black] (6.5,-0.3) -- (6.5,-0.78);
\node[yearbox] at (6.5,0) {2022};

\draw[line width=0.6pt, blue!50!black] (8.125,0.3) -- (8.125,0.78);
\draw[line width=0.6pt, purple!60!black] (8.125,-0.3) -- (8.125,-0.78);
\node[yearbox] at (8.125,0) {2023};

\draw[line width=0.6pt, blue!50!black] (9.75,0.3) -- (9.75,0.78);
\draw[line width=0.6pt, purple!60!black] (9.75,-0.3) -- (9.75,-0.78);
\node[yearbox] at (9.75,0) {2024};

\draw[line width=0.6pt, blue!50!black] (11.375,0.3) -- (11.375,0.78);
\draw[line width=0.6pt, purple!60!black] (11.375,-0.3) -- (11.375,-0.78);
\node[yearbox] at (11.375,0) {2025};

\draw[line width=0.6pt, blue!50!black] (13,0.3) -- (13,0.78);
\draw[line width=0.6pt, purple!60!black] (13,-0.3) -- (13,-0.78);
\node[yearbox] at (13,0) {2026};

\node[approach=blue!8, text width=0.8cm] at (0,1.4) {\cite{Lalitha2018FullyDF}};

\node[approach=blue!8, text width=0.8cm] at (1.625,1.4) {\cite{DBLP:journals/corr/abs-1905-06731}};
\node[approach=blue!8, text width=0.8cm] at (1.625,1.8) {\cite{hu2019decentralized}};


\node[approach=blue!8, text width=0.8cm] at (4.875,1.4) {\cite{9472790}};
\node[approach=blue!8, text width=0.8cm] at (4.875,1.8) {\cite{9687521}};

\node[approach=blue!8, text width=0.8cm] at (6.5,1.4) {\cite{9426904}};
\node[approach=blue!8, text width=0.8cm] at (6.5,1.8) {\cite{9777682}};
\node[approach=blue!8, text width=0.8cm] at (6.5,2.2) {\cite{9700624}};
\node[approach=blue!8, text width=0.8cm] at (6.5,2.6) {\cite{9825726}};
\node[approach=blue!8, text width=0.8cm] at (6.5,3.0) {\cite{li2022panm}};

\node[approach=blue!8, text width=0.8cm] at (8.125,1.4) {\cite{kalra2023decentralized}};
\node[approach=blue!8, text width=0.8cm] at (8.125,1.8) {\cite{9996127}};

\node[approach=blue!8, text width=0.8cm] at (9.75,1.4) {\cite{chai2024df-svd}};
\node[approach=blue!8, text width=0.8cm] at (9.75,1.8) {\cite{gu2024statisticalinferencefordfl}};
\node[approach=blue!8, text width=0.8cm] at (9.75,2.2) {\cite{fang2024byzantine-dfl}};
\node[approach=blue!8, text width=0.8cm] at (9.75,2.6) {\cite{huang2024modelheterogeneity-dfl}};

\node[approach=blue!8, text width=0.8cm] at (11.375,1.4) {\cite{jia2025selfishattack}};
\node[approach=blue!8, text width=0.8cm] at (11.375,1.8) {\cite{zehtabi2025dfl-convergenceguarantees}};
\node[approach=blue!8, text width=0.8cm] at (11.375,2.2) {\cite{thompson2025ntkdfl}};
\node[approach=blue!8, text width=0.8cm] at (11.375,2.6) {\cite{li2025mitigating}};
\node[approach=blue!8, text width=0.8cm] at (11.375,3) {\cite{bonnerjee2025sharp}};

\node[approach=blue!8, text width=0.8cm] at (13,1.4) {\cite{fenoglio2025ERISEnhancingPrivacy}};
\node[approach=blue!8, text width=0.8cm] at (13,1.8) {\cite{11230822}};

\node[approach=purple!8, text width=0.8cm] at (1.625,-1.4) {\cite{zhou2019pirate}};

\node[approach=purple!8, text width=0.8cm] at (3.25,-1.4) {\cite{9182705}};
\node[approach=purple!8, text width=0.8cm] at (3.25,-1.8) {\cite{8733825}};
\node[approach=purple!8, text width=0.8cm] at (3.25,-2.2) {\cite{liu2020fedcoin}};

\node[approach=purple!8, text width=0.8cm] at (4.875,-1.4) {\cite{hu2021gfl}};
\node[approach=purple!8, text width=0.8cm] at (4.875,-1.8) {\cite{9170559}};
\node[approach=purple!8, text width=0.8cm] at (4.875,-2.2) {\cite{9382023}};
\node[approach=purple!8, text width=0.8cm] at (4.875,-2.6) {\cite{8894364}};
\node[approach=purple!8, text width=0.8cm] at (4.875,-3.0) {\cite{9292450}};

\node[approach=purple!8, text width=0.8cm] at (6.5,-1.4) {\cite{9399813}};
\node[approach=purple!8, text width=0.8cm] at (6.5,-1.8) {\cite{9664296}};

\node[approach=purple!8, text width=0.8cm] at (8.125,-1.4) {\cite{9866512}};

\end{tikzpicture}
\caption{Timeline of the works included in our taxonomy, grouped by publication year and categorized as traditional DFL (top) versus blockchain-based FL (bottom). The distribution highlights that blockchain-based proposals are concentrated mainly in 2019--2022 within our corpus, whereas traditional DFL continues to attract sustained research activity in later years (2023--2025), reflecting a shift of emphasis toward lighter-weight P2P coordination and direct mechanisms for robustness, heterogeneity, privacy, and communication efficiency.}
\label{fig:timeline}
\end{figure*}
Several attempts have been made to overcome the limitations posed by the centralized orchestration of standard FL.
In this section, we present the most relevant approaches toward DFL proposed in the literature. 

We, first, broadly categorize existing methods into the following two-class taxonomy:
\begin{itemize}
\item \textit{\textbf{Traditional Distributed (TD-FL):}} Methods in this class obtain FL decentralization through ``standard'' distributed computing techniques.
\item \textit{\textbf{Blockchain-based (BC-FL):}} This class contains all the approaches that achieve FL decentralization by exploiting blockchain  functionalities.
\end{itemize}
Then, we segment the literature into the main objectives which the studies aim to tackle (Table~\ref{tab:taxonomy}).


Figure~\ref{fig:timeline} provides a chronological view of the works included in our taxonomy, separated into traditional DFL and blockchain-based FL. A clear pattern emerges: BC-FL is concentrated in a relatively narrow window (roughly 2019–2022), after which the stream of novel proposals in our corpus largely disappears, while TD-FL continues to grow through 2024–2025 with multiple new directions.

This does not necessarily mean that blockchain-based FL has “failed”, but it does suggest that it has not remained an active primary research thrust for DFL. A plausible interpretation is that, once the community clarified the trade-offs, blockchains were increasingly treated as an enabling component for specific requirements (e.g., auditability, accountability, incentive settlement) rather than as a default substrate for decentralization. In many settings, the additional overheads introduced by the blockchain (see Section~\ref{subsec:Blockchain})
can be hard to justify compared to lighter-weight P2P coordination, especially when the core FL bottlenecks are communication and system heterogeneity rather than the absence of a ledger. As a result, research attention appears to have shifted toward more direct solutions to robustness, heterogeneity, privacy, and communication efficiency in fully decentralized settings, which remain open and evolving.

Table~\ref{tab:taxonomy} provides a compact cross-paper view of which challenges are explicitly targeted in the DFL literature. A clear pattern is that TD-FL works increasingly emphasize communication/bandwidth efficiency and heterogeneity handling, often accompanied by stronger theoretical analyses, whereas BC-FL works more frequently motivate security and auditability goals. Overall, the table highlights that only a subset of approaches address multiple dimensions simultaneously, suggesting that designing DFL systems that are secure, privacy-preserving, robust to faults/adversaries, and communication-efficient remains largely an open problem.
\begin{center}
\begin{table*}[!htb]
    \centering
    \begin{adjustbox}{width=1\textwidth}
    \rowcolors{2}{gray!10}{white}
    \begin{tabular}{l | l c c c c c c c c}
    \toprule
     \textbf{Category} & \textbf{Work} &
     \textbf{Fault Tolerance} &
     \textbf{Privacy} &
     \textbf{Security} &
     \textbf{\thead{Incentive\\Mech.}} &
     \textbf{\thead{Data\\Heterog.}} &
     \textbf{\thead{Bandwidth\\Utiliz.}} &
     \textbf{\thead{Theory\\\& Task Ext.}} &
     \textbf{\thead{Resource\\Eff.}} \\
     \midrule
     
     & IPLS \cite{9472790}
       & \CIRCLE & \Circle & \Circle & \Circle & \Circle & \Circle & \Circle & \LEFTcircle \\
     & DFL-UN \cite{9687521}
       & \CIRCLE & \Circle & \Circle & \Circle & \Circle & \Circle & \Circle & \Circle \\
     & DSpodFL~\cite{zehtabi2025dfl-convergenceguarantees}
       & \CIRCLE & \Circle & \Circle & \Circle & \LEFTcircle & \Circle & \LEFTcircle & \Circle \\
     & ProxyFL \cite{kalra2023decentralized}
       & \Circle & \CIRCLE & \Circle & \Circle & \Circle & \LEFTcircle & \Circle & \Circle \\
     & PVD-FL \cite{9777682}
       & \Circle & \CIRCLE & \LEFTcircle & \Circle & \Circle & \Circle & \Circle & \Circle \\
     & f-Differential Privacy~\cite{li2025mitigating}
       & \Circle & \CIRCLE & \Circle & \Circle & \Circle & \Circle & \LEFTcircle & \Circle \\
     & ERIS \cite{fenoglio2025ERISEnhancingPrivacy}
       & \Circle & \CIRCLE & \Circle & \Circle & \Circle & \LEFTcircle & \Circle & \LEFTcircle \\
     & Trusted DFL \cite{9700624}
       & \Circle & \Circle & \CIRCLE & \Circle & \Circle & \Circle & \Circle & \Circle \\
     & BALANCE~\cite{fang2024byzantine-dfl}
       & \Circle & \Circle & \CIRCLE & \Circle & \Circle & \Circle & \LEFTcircle & \Circle \\
    & BR-DFL \cite{11230822} 
       & \Circle & \Circle & \CIRCLE & \Circle & \CIRCLE & \LEFTcircle & \Circle & \Circle \\
     & SelfishAttack~\cite{jia2025selfishattack}
       & \Circle & \Circle & \CIRCLE & \Circle & \Circle & \Circle & \Circle & \Circle \\
     & Combo \cite{hu2019decentralized}
       & \Circle & \Circle & \Circle & \Circle & \Circle & \CIRCLE & \Circle & \Circle \\
     & GossipFL \cite{9996127}
       & \Circle & \Circle & \Circle & \Circle & \Circle & \CIRCLE & \Circle & \Circle \\
     & Def-KT \cite{9426904}
       & \Circle & \Circle & \Circle & \Circle & \CIRCLE & \Circle & \Circle & \Circle \\
     & P2P FL over graphs \cite{9825726}
       & \Circle & \Circle & \Circle & \Circle & \CIRCLE & \Circle & \LEFTcircle & \Circle \\
     & PANM \cite{li2022panm}
       & \Circle & \Circle & \Circle & \Circle & \CIRCLE & \LEFTcircle & \LEFTcircle & \Circle \\
     & DeSA~\cite{huang2024modelheterogeneity-dfl}
       & \Circle & \Circle & \Circle & \Circle & \CIRCLE & \Circle & \Circle & \Circle \\
     & NTK-DFL~\cite{thompson2025ntkdfl}
       & \Circle & \Circle & \Circle & \Circle & \CIRCLE & \LEFTcircle & \Circle & \Circle \\
     & FDFL \cite{Lalitha2018FullyDF}
       & \Circle & \Circle & \Circle & \Circle & \Circle & \Circle & \CIRCLE & \Circle \\
     & BrainTorrent \cite{DBLP:journals/corr/abs-1905-06731}
       & \LEFTcircle & \Circle & \Circle & \Circle & \Circle & \Circle & \CIRCLE & \Circle \\
     & Sharp Gaussian approximations~\cite{bonnerjee2025sharp}
       & \Circle & \Circle & \Circle & \Circle & \Circle & \Circle & \CIRCLE & \Circle \\
     & Statistical inference for DFL~\cite{gu2024statisticalinferencefordfl}
       & \Circle & \Circle & \Circle & \Circle & \LEFTcircle & \Circle & \CIRCLE & \Circle \\
     \multirow{-23}{*}{\textbf{TD-FL}} & Excalibur~\cite{chai2024df-svd}
       & \Circle & \LEFTcircle & \Circle & \Circle & \Circle & \LEFTcircle & \CIRCLE & \LEFTcircle \\

     \midrule

     & FED-BC \cite{9182705}
       & \CIRCLE & \Circle & \Circle & \Circle & \Circle & \Circle & \Circle & \Circle \\
     & GFL \cite{hu2021gfl}
       & \CIRCLE & \Circle & \LEFTcircle & \Circle & \Circle & \Circle & \Circle & \Circle \\
     & Privacy-Preserving for IoT Devices \cite{9170559}
       & \Circle & \CIRCLE & \LEFTcircle & \Circle & \Circle & \Circle & \Circle & \Circle \\
     & DeepChain \cite{8894364}
       & \Circle & \CIRCLE & \CIRCLE & \LEFTcircle & \Circle & \Circle & \Circle & \Circle \\
     & Biscotti \cite{9292450}
       & \Circle & \CIRCLE & \CIRCLE & \Circle & \Circle & \Circle & \Circle & \Circle \\
     & BytoChain \cite{9382023}
       & \Circle & \Circle & \CIRCLE & \Circle & \Circle & \Circle & \Circle & \Circle \\
     & PIRATE \cite{zhou2019pirate}
       & \Circle & \Circle & \CIRCLE & \Circle & \Circle & \Circle & \Circle & \Circle \\
     & Trustworthy FL \cite{9866512}
       & \Circle & \Circle & \CIRCLE & \Circle & \Circle & \Circle & \Circle & \Circle \\
     & BlockFL \cite{8733825}
       & \Circle & \Circle & \LEFTcircle & \CIRCLE & \Circle & \Circle & \Circle & \LEFTcircle \\
     & FedCoin \cite{liu2020fedcoin}
       & \Circle & \Circle & \Circle & \CIRCLE & \Circle & \Circle & \Circle & \Circle \\
     & BAFL \cite{9399813}
       & \Circle & \Circle & \LEFTcircle & \LEFTcircle & \Circle & \Circle & \Circle & \CIRCLE \\
     \multirow{-11}{*}{\textbf{BC-FL}} & BLADE-FL \cite{9664296}
       & \Circle & \LEFTcircle & \Circle & \LEFTcircle & \Circle & \Circle & \Circle & \CIRCLE \\

     \bottomrule
     \end{tabular}
     \end{adjustbox}
    \caption{This table summarizes the main challenges addressed by the DFL works included in this survey. Each row corresponds to a method, grouped by traditional DFL (TD-FL) and blockchain-assisted DFL (BC-FL); columns indicate whether the paper explicitly tackles (\CIRCLE) or partially considers (\LEFTcircle) key dimensions—fault tolerance, privacy, security, incentive mechanisms, data heterogeneity, bandwidth utilization, theoretical/extension aspects, and resource efficiency.}
    \label{tab:taxonomy}
\end{table*}
\end{center}

Table~\ref{tab:experimental-settings} highlights a strong concentration of empirical evidence around a small set of benchmark datasets and relatively standard model families. Across both TD-FL and BC-FL, most evaluations rely on horizontal partitioning, with vertical FL appearing only sporadically, and a recurring use of MNIST/FMNIST/CIFAR-style benchmarks. This evidence may overestimate practical readiness, since many studies are validated on small, clean benchmarks rather than realistic, heterogeneous, resource-constrained deployments. Overall, the table motivates a need for more standardized and transparent evaluation protocols for DFL, including consistent reporting of partitioning schemes and broader coverage of real-world datasets and models.
\begin{center}
\begin{table*}[!htb]
    \centering
    \begin{adjustbox}{width=1\textwidth}
    \rowcolors{2}{gray!10}{white}
    \begin{tabular}{l|llll}
    \toprule
     \textbf{Category} & \textbf{Work} & \textbf{Model} & \textbf{Data Partitioning} & \textbf{Dataset} \\
     \midrule
     \multirow{22}{*}{TD-FL} & IPLS \cite{9472790} & NN & Vertical, IID & MNIST \\
     & DFL-UN \cite{9687521} & CNN & Horizontal, Non-IID & N/A \\
     & DSpodFL \cite{zehtabi2025dfl-convergenceguarantees} & SVM, VGG11 & Horizontal, IID, Non-IID & CIFAR-10, FMNIST \\
     & ProxyFL \cite{kalra2023decentralized} & NN, CNN & Horizontal, IID, Non-IID & MNIST, FMNIST, CIFAR-10 \\
     & PVD-FL \cite{9777682} & NN & Vertical, IID & MNIST, Thyroid, Breast Cancer, German Credit \\
     & f-Differential Privacy \cite{li2025mitigating} & Logistic Regression, CNN & Horizontal, Non-IID & UCI housing, MNIST, a9a, synthetic \\
     & ERIS \cite{fenoglio2025ERISEnhancingPrivacy} & GPT-Neo, DistilBERT, ResNet-9, LeNet-5 & Horizontal, IID, Non-IID & CNN/DailyMail, IMDB, CIFAR-10, MNIST \\
     & Trusted DFL \cite{9700624} & NN & Horizontal, Non-IID & N/A \\
     & BALANCE \cite{fang2024byzantine-dfl} & CNN & Horizontal, Non-IID & MNIST, FMNIST, HAR, CelebA \\
     & BR-DFL \cite{11230822} & ResNet-18 & Horizontal, IID, Non-IID & GTSRB, CIFAR-10, EMNIST \\ 
     & SelfishAttack \cite{jia2025selfishattack} & CNN, LSTM & Horizontal, Non-IID & CIFAR-10, FEMNIST, Sent140 \\
     & Combo \cite{hu2019decentralized} & CNN & Horizontal, IID & CIFAR-10 \\
     & GossipFL \cite{9996127} & CNN, ResNet-20 & Horizontal, IID, Non-IID & MNIST, CIFAR-10 \\
     & Def-KT \cite{9426904} & NN, CNN & Horizontal, IID, Non-IID & MNIST, FMNIST, CIFAR-10, CIFAR-100 \\
     & P2P FL over graphs \cite{9825726} & Bayesian Linear Regression & Horizontal, Non-IID & MNIST, FMNIST, FEMNIST \\
     & PANM \cite{li2022panm} & NN, CNN & Horizontal, Non-IID & MNIST, FMNIST, CIFAR-10 \\
     & DeSA \cite{huang2024modelheterogeneity-dfl} & AlexNet, CNN & Horizontal, Non-IID & MNIST, SVHN, USPS, SynthDigits, MNIST-M, Amazon, Caltech, DSLR, Webcam, CIFAR10C \\
     & NTK-DFL \cite{thompson2025ntkdfl} & MLP & Horizontal, IID, Non-IID & MNIST, FMNIST, FEMNIST \\
     & FDFL \cite{Lalitha2018FullyDF} & N/A & Horizontal, Non-IID & N/A \\
     & BrainTorrent \cite{DBLP:journals/corr/abs-1905-06731} & Quicknat & Horizontal, Non-IID & MALC \\
     & Sharp Gaussian approximations \cite{bonnerjee2025sharp} & Linear Classifier & Horizontal, Non-IID & MNIST \\
     & Statistical inference for DFL \cite{gu2024statisticalinferencefordfl} & Linear Regression, Logistic Regression & Horizontal, Non-IID &  Synthetic \\
     \multirow{-23}{*}{\textbf{TD-FL}} & Excalibur \cite{chai2024df-svd} & SVD & Vertical, Non-IID & MNIST, Wine, ML100K, synthetic \\
     \midrule
     \multirow{12}{*}{BC-FL} & FED-BC \cite{9182705} & NN & Horizontal, Non-IID & MNIST \\
     & GFL \cite{hu2021gfl} & CNN & Horizontal, Non-IID & MNIST, CIFAR-10, CIFAR-100 \\
     & Privacy-Preserving for IoT Devices \cite{9170559} & CNN & Horizontal, IID & MNIST \\
     & DeepChain \cite{8894364} & CNN & Horizontal, IID & MNIST \\ 
     & Biscotti \cite{9292450} & Linear Regression, Softmax Classifier & Horizontal, IID & Credit Card Fraud, MNIST \\
     & BytoChain \cite{9382023} & CNN & Horizontal, IID & MNIST \\
     & PIRATE \cite{zhou2019pirate} & NN & Horizontal, IID & N/A \\
     & Trustworthy FL \cite{9866512} & NN, CNN, AlexNet & Horizontal, IID, Non-IID & MNIST, CIFAR-10, HeartActivity \\
     & BlockFL \cite{8733825} & NN & Horizontal, IID & N/A \\
     & FedCoin \cite{liu2020fedcoin} & N/A & Horizontal, IID & MNIST \\
     & BAFL \cite{9399813} & CNN & Horizontal, IID & MNIST \\
     \multirow{-11}{*}{\textbf{BC-FL}} & BLADE-FL \cite{9664296} & NN, VGG$_{11}$ & Horizontal, Non-IID & MNIST, CIFAR-10, FMNIST, SVHN \\
     \bottomrule
     \end{tabular}
     \end{adjustbox}
    \caption{This table reports the experimental setups used by the DFL works surveyed in this paper. For each method (grouped into TD-FL and BC-FL), we list the model family, the data partitioning setting (horizontal/vertical; IID/non-IID), and the datasets adopted in the empirical evaluation.}
    \label{tab:experimental-settings}
\end{table*}
\end{center}

Figure~\ref{fig:taxonomy} links the survey organization to the literature: each column corresponds to a challenge dimension (and the section where it is discussed), and the listed papers are representative works whose \emph{primary} contribution targets that dimension. The color split (TD-FL vs.\ BC-FL) highlights a qualitative specialization: BC-FL concentrates, in proportion, on security and incentive mechanisms where a ledger can provide auditability, whereas TD-FL spans a broader set of system-centric problems—especially heterogeneity handling, communication efficiency, and theory/task extensions—that have continued to evolve in the most recent years. This mapping also exposes gaps in coverage, which we revisit as open problems later in the survey.
\begin{figure*}[!htb]
\centering
\scalebox{0.85}{
\begin{tikzpicture}[
    category/.style={
        rectangle,
        rounded corners=3pt,
        minimum width=2.8cm,
        minimum height=1.2cm,
        align=center,
        draw=black,
        line width=0.8pt,
        font=\sffamily\small
    },
    traditional/.style={
        rectangle,
        rounded corners=2pt,
        minimum width=2.2cm,
        minimum height=0.5cm,
        align=center,
        draw=blue!40!black,
        fill=blue!8,
        line width=0.6pt,
        font=\sffamily\footnotesize
    },
    blockchain/.style={
        rectangle,
        rounded corners=2pt,
        minimum width=2.2cm,
        minimum height=0.5cm,
        align=center,
        draw=purple!40!black,
        fill=purple!8,
        line width=0.6pt,
        font=\sffamily\footnotesize
    },
    connection/.style={
        draw=black!70,
        line width=1.2pt
    }
]

\node[category] (fault) at (0,0) {Fault Tolerance};
\node[category] (privacy) at (4,0) {Privacy};
\node[category] (security) at (8,0) {Security};
\node[category] (incentive) at (12,0) {Incentive\\Mechanisms};

\node[category] (data) at (0,-10) {Data Heterogeneity};
\node[category] (bandwidth) at (4,-10) {Bandwidth\\Utilization};
\node[category] (theory) at (8,-10) {Theory \& Task Ext.};
\node[category] (resource) at (12,-10) {Resource Effic.};

\node[traditional] (ipls) at (0.5,-1.5) {\hyperref[paragraph:ipls]{IPLS}~\cite{9472790}};
\node[traditional] (dfl) at (0.5,-2.5) {\hyperref[paragraph:dfl-un]{DFL-UN}~\cite{flUAV}};
\node[traditional] (dspod) at (0.5,-3.5) {\hyperref[paragraph:dspodfl]{DSpodFL}~\cite{zehtabi2025dfl-convergenceguarantees}};
\node[blockchain] (fed) at (0.5,-4.5) {\hyperref[paragraph:fed-bc]{FED-BC}~\cite{9182705}};
\node[blockchain] (gfl) at (0.5,-5.5) {\hyperref[paragraph:gfl]{GFL}~\cite{hu2021gfl}};

\node[traditional] (proxy) at (4.5,-1.5) {\hyperref[paragraph:proxyfl]{ProxyFL}~\cite{kalra2023decentralized}};
\node[traditional] (pvd) at (4.5,-2.5) {\hyperref[paragraph:pvd-fl]{PVD-FL}~\cite{9777682}};
\node[traditional] (diff) at (4.5,-3.5) {\hyperref[paragraph:f-differentialprivacy]{f-DifferentialPrivacy}~\cite{li2025mitigating}};
\node[traditional] (eris) at (4.5,-4.5) {\hyperref[paragraph:eris]{ERIS}~\cite{fenoglio2025ERISEnhancingPrivacy}};
\node[blockchain] (iot) at (4.5,-5.5) {\hyperref[paragraph:privacy-preservingflforiotdevices]{Privacy Preserving for}\\\hyperref[paragraph:privacy-preservingflforiotdevices]{IoT Devices}~\cite{9170559}};
\node[blockchain] (deep) at (4.5,-6.5) {\hyperref[paragraph:deepchain]{DeepChain}~\cite{8894364}};
\node[blockchain] (biscotti) at (4.5,-7.5) {\hyperref[paragraph:biscotti]{Biscotti}~\cite{9292450}};

\node[traditional] (trusted) at (8.5,-1.5) {\hyperref[paragraph:trusted-dfl]{Trusted DFL}~\cite{9700624}};
\node[traditional] (balance) at (8.5,-2.5) {\hyperref[paragraph:balance]{BALANCE}~\cite{fang2024byzantine-dfl}};
\node[traditional] (backdoor_resilient) at (8.5,-3.5) {\hyperref[paragraph:brdfl]{BR-DFL} \cite{11230822}};
\node[traditional] (selfish) at (8.5,-4.5) {\hyperref[paragraph:selfishattack]{SelfishAttack}~\cite{jia2025selfishattack}};
\node[blockchain] (bytochain) at (8.5,-5.5) {\hyperref[paragraph:bytochain]{BytoChain}~\cite{9382023}};
\node[blockchain] (pirate) at (8.5,-6.5) {\hyperref[paragraph:pirate]{PIRATE}~\cite{zhou2019pirate}};
\node[blockchain] (trustworthy) at (8.5,-7.5) {\hyperref[paragraph:trustworthyfl]{Trustworthy FL} \cite{9866512}};

\node[blockchain] (inc1) at (12.5,-1.5) {\hyperref[paragraph:blockfl]{BlockFL} \cite{8733825}};
\node[blockchain] (inc2) at (12.5,-2.5) {\hyperref[paragraph:fedcoin]{FedCoin} \cite{liu2020fedcoin}};

\node[traditional] (defkt) at (0.5,-11.5) {\hyperref[paragraph:def-kt]{Def-KT}~\cite{9426904}};
\node[traditional] (p2pfl) at (0.5,-12.5) {\hyperref[paragraph:p2p-fl-over-graphs]{P2P FL over graphs}~\cite{9825726}};
\node[traditional] (panm) at (0.5,-13.5) {\hyperref[paragraph:panm]{PANM}~\cite{li2022panm}};
\node[traditional] (desa) at (0.5,-14.5) {\hyperref[paragraph:desa]{DeSA}~\cite{huang2024modelheterogeneity-dfl}};
\node[traditional] (ntk) at (0.5,-15.5) {\hyperref[paragraph:ntk-dfl]{NTK-DFL}~\cite{thompson2025ntkdfl}};

\node[traditional] (combo) at (4.5,-11.5) {\hyperref[paragraph:combo]{Combo} \cite{hu2019decentralized}};
\node[traditional] (gossip) at (4.5,-12.5) {\hyperref[paragraph:gossipfl]{GossipFL} \cite{9996127}};

\node[traditional] (fdfl) at (8.5,-11.5) {\hyperref[paragraph:fdfl]{FDFL}~\cite{Lalitha2018FullyDF}};
\node[traditional] (braintorrent) at (8.5,-12.5) {\hyperref[paragraph:braintorrent]{BrainTorrent}~\cite{DBLP:journals/corr/abs-1905-06731}};
\node[traditional] (sharp) at (8.5,-13.5) {\hyperref[paragraph:sharpgaussianapproximations]{Sharp Gaussian} \\\hyperref[paragraph:sharpgaussianapproximations]{approximations}~\cite{bonnerjee2025sharp}};
\node[traditional] (statinference) at (8.5,-14.5) {\hyperref[paragraph:statisticalinferencefordfl]{Statistical inference}\\ \hyperref[paragraph:statisticalinferencefordfl]{for DFL}~\cite{gu2024statisticalinferencefordfl}};
\node[traditional] (excalibur) at (8.5,-15.5) {\hyperref[paragraph:excalibur]{Excalibur}~\cite{chai2024df-svd}};

\node[blockchain] (bafl) at (12.5,-11.5) {\hyperref[paragraph:bafl]{BAFL} \cite{9399813}};
\node[blockchain] (blade) at (12.5,-12.5) {\hyperref[paragraph:blade-fl]{BLADE-FL} \cite{9664296}};

\coordinate (fvline) at ($(fault.south west)+(0.15,0)$);
\draw[connection] (fvline) -- ++(0,-4.92);
\draw[connection] ($(fvline)+(0,-0.9)$) -| (ipls.west);
\draw[connection] ($(fvline)+(0,-1.9)$) -| (dfl.west);
\draw[connection] ($(fvline)+(0,-2.9)$) -| (dspod.west);
\draw[connection] ($(fvline)+(0,-3.9)$) -| (fed.west);
\draw[connection] ($(fvline)+(0,-4.9)$) -| (gfl.west);

\coordinate (pvline) at ($(privacy.south west)+(0.15,0)$);
\draw[connection] (pvline) -- ++(0,-6.92);
\draw[connection] ($(pvline)+(0,-0.9)$) -| (proxy.west);
\draw[connection] ($(pvline)+(0,-1.9)$) -| (pvd.west);
\draw[connection] ($(pvline)+(0,-2.9)$) -| (diff.west);
\draw[connection] ($(pvline)+(0,-3.9)$) -| (eris.west);
\draw[connection] ($(pvline)+(0,-4.9)$) -| (iot.west);
\draw[connection] ($(pvline)+(0,-5.9)$) -| (deep.west);
\draw[connection] ($(pvline)+(0,-6.9)$) -| (biscotti.west);

\coordinate (svline) at ($(security.south west)+(0.15,0)$);
\draw[connection] (svline) -- ++(0,-6.92);
\draw[connection] ($(svline)+(0,-0.9)$) -| (trusted.west);
\draw[connection] ($(svline)+(0,-1.9)$) -| (balance.west);
\draw[connection] ($(svline)+(0,-2.9)$) -| (backdoor_resilient.west);
\draw[connection] ($(svline)+(0,-3.9)$) -| (selfish.west);
\draw[connection] ($(svline)+(0,-4.9)$) -| (bytochain.west);
\draw[connection] ($(svline)+(0,-5.9)$) -| (pirate.west);
\draw[connection] ($(svline)+(0,-6.9)$) -| (trustworthy.west);

\coordinate (ivline) at ($(incentive.south west)+(0.15,0)$);
\draw[connection] (ivline) -- ++(0,-1.92);
\draw[connection] ($(ivline)+(0,-0.9)$) -| (inc1.west);
\draw[connection] ($(ivline)+(0,-1.9)$) -| (inc2.west);

\coordinate (dvline) at ($(data.south west)+(0.15,0)$);
\draw[connection] (dvline) -- ++(0,-4.92);
\draw[connection] ($(dvline)+(0,-0.9)$) -| (defkt.west);
\draw[connection] ($(dvline)+(0,-1.9)$) -| (p2pfl.west);
\draw[connection] ($(dvline)+(0,-2.9)$) -| (panm.west);
\draw[connection] ($(dvline)+(0,-3.9)$) -| (desa.west);
\draw[connection] ($(dvline)+(0,-4.9)$) -| (ntk.west);

\coordinate (bvline) at ($(bandwidth.south west)+(0.15,0)$);
\draw[connection] (bvline) -- ++(0,-1.92);
\draw[connection] ($(bvline)+(0,-0.9)$) -| (combo.west);
\draw[connection] ($(bvline)+(0,-1.9)$) -| (gossip.west);

\coordinate (tvline) at ($(theory.south west)+(0.15,0)$);
\draw[connection] (tvline) -- ++(0,-4.92);
\draw[connection] ($(tvline)+(0,-0.9)$) -| (fdfl.west);
\draw[connection] ($(tvline)+(0,-1.9)$) -| (braintorrent.west);
\draw[connection] ($(tvline)+(0,-2.9)$) -| (sharp.west);
\draw[connection] ($(tvline)+(0,-3.9)$) -| (statinference.west);
\draw[connection] ($(tvline)+(0,-4.9)$) -| (excalibur.west);

\coordinate (rvline) at ($(resource.south west)+(0.15,0)$);
\draw[connection] (rvline) -- ++(0,-1.92);
\draw[connection] ($(rvline)+(0,-0.9)$) -| (bafl.west);
\draw[connection] ($(rvline)+(0,-1.9)$) -| (blade.west);
\end{tikzpicture}
}
\caption{Mapping between the survey structure and the taxonomy in Table~\ref{tab:taxonomy}. Each column corresponds to (i) a main challenge dimension in DFL and (ii) the dedicated section where that dimension is discussed in the survey. Within each column, blue boxes denote TD-FL works and red boxes denote BC-FL works, listing representative papers that primarily address that challenge.}
\label{fig:taxonomy}
\end{figure*}

\subsection{Traditional Distributed FL (TD-FL)}
A possible solution to overcome the limitations of centralized FL is to adopt decentralized frameworks inspired by traditional distributed computing technologies, such as P2P networks.


In this work, we classify existing contributions falling into this category into six groups: \textit{fault-tolerance}, \textit{privacy and security}, \textit{data heterogeneity}, \textit{bandwidth utilization}, and \textit{theory/task extensions}. These categories represent the authors' main contributions; however, it is important to note that an article categorized into one group may also solve (in part) issues belonging to other groups, as shown in Table~\ref{tab:taxonomy}. 

\subsubsection{\underline{Fault-tolerance}}
\label{subsubsec:tdfl-fault-tolerance}
In DFL, fault tolerance is mainly challenged by churn (nodes joining/leaving), intermittent or unreliable links, and the absence of a stable coordinator that can enforce participation, retransmissions, or global synchronization. TD-FL works tackle these issues by designing decentralized mechanisms that keep model information flowing despite partial connectivity: some rely on P2P dissemination layers that make model states retrievable even when peers disconnect, while others use neighbor-based aggregation and repeated broadcasts so updates can propagate opportunistically through the network. The works below exemplify these directions in different environments, from generic P2P overlays to highly dynamic UAV swarms, and from system-oriented designs to algorithmic models of sporadic computation and communication.

\phantomsection \textbf{\textit{IPLS~\cite{9472790}.}} \label{paragraph:ipls}
IPLS is a fully decentralized DFL framework built on interplanetary file system~\cite{DBLP:journals/corr/Benet14} (IPFS), designed to support open participation and robustness to churn and intermittent connectivity. Nodes fetch and publish model artifacts through the P2P storage layer, enabling training to progress without a fixed coordinator; experimentally, accuracy approaches centralized baselines with minimal loss under the considered settings. Its limitations are mostly system-level: the evaluation is still narrow (device heterogeneity and large modern models are not explored) and the approach inherits practical overheads/assumptions from the underlying content-addressed networking layer.

\phantomsection \textbf{\textit{DFL-UN~\cite{9687521}.}} \label{paragraph:dfl-un}
DFL-UN studies decentralized FL in UAV networks, where links are unreliable and topology changes quickly. Each UAV trains locally, aggregates models from one-hop neighbors, and broadcasts the updated model, avoiding a single point of failure and adapting naturally to link disruptions. The main limitations are that robustness/convergence is evaluated in small-scale settings and the design must cope with real UAV constraints (rapid mobility, unstable wireless channels, and transient connectivity) that can hinder stable convergence and timely model propagation.
%

\phantomsection \textbf{\textit{DSpodFL~\cite{zehtabi2025dfl-convergenceguarantees}.}} \label{paragraph:dspodfl}
DSpodFL targets intermittent links between nodes in DFL by modeling that peers may not compute or communicate at every step. It introduces a unified update rule where local training and neighbor exchanges are governed by indicator variables (capturing sporadic computation/communication), so several classical decentralized methods emerge as special cases. The paper provides convergence guarantees and shows improved time-to-accuracy in heterogeneous settings.

Main limitations are that the sporadicity processes are abstracted as indicator variables (so the practical benefit depends on how well they match real scheduling/link dynamics), and the empirical validation is limited to a small set of vision tasks/models (FMNIST/CIFAR10 with relatively standard classifiers), leaving broader workloads and deployment-driven constraints for follow-up work.
%

\subsubsection{\underline{Privacy}}
\label{subsubsec:tdfl-privacy}
In DFL, nodes must exchange models/updates to coordinate, which can leak information about local data even when raw data never leave devices. TD-FL privacy mechanisms therefore aim to reduce what each message reveals, mainly via (i) proxy/public representations (ProxyFL~\cite{kalra2023decentralized}), (ii) cryptographic confidentiality with verifiability (PVD-FL~\cite{9777682}), (iii) decentralized DP accounting/noise tailored to networked communication (f-DP~\cite{li2025mitigating}), and (iv) update sharding across multiple aggregators (often coupled with compression) to limit any single observer’s view (ERIS~\cite{fenoglio2025ERISEnhancingPrivacy}).

\phantomsection \textbf{\textit{ProxyFL~\cite{kalra2023decentralized}.}} \label{paragraph:proxyfl}
In this scheme, participants maintain two separate models: a private model and a publicly shared proxy model. The purpose of the proxy model is to protect the privacy of individual participants by not sharing the private model while also enabling efficient information exchange without the need for a centralized server.
Key advantages of ProxyFL include reduced communication overhead and enhanced privacy protection. Participants can have their private models with different architectures, promoting model heterogeneity. Experimental results on image datasets and a cancer diagnostic problem show that ProxyFL outperforms existing alternatives in terms of communication efficiency and convergence speed.

\phantomsection \textbf{\textit{PVD-FL~\cite{9777682}.}} \label{paragraph:pvd-fl}
PVD-FL primarily addresses privacy and verifiability in decentralized FL: it aims to keep both the global model and local updates confidential during training, while allowing peers to verify that training computations are carried out correctly (mitigating faulty/lazy behavior) without relying on a trusted server. Methodologically, the framework is built on an efficient verifiable cipher-based matrix multiplication primitive, which enables encrypted linear-algebra operations with correctness checks and is then used to implement decentralized training steps.
Its main limitations are in scope and assumptions: the design focuses on confidentiality and correctness of computation, but does not directly handle data/model poisoning (acknowledged as a future direction), and it relies on simplifying system settings (e.g., non-collusion and IID data) while incurring cryptographic overhead.

\phantomsection \textbf{\textit{f-Differential Privacy~\cite{li2025mitigating}.}} \label{paragraph:f-differentialprivacy}
The paper investigates privacy accounting methods for differentially private DFL. They focus on two DFL algorithms within the f-DP framework and propose two new ways of quantifying privacy leakage that better support decentralized communication and noise structure.

The authors introduce two f-DP notions tailored to decentralized settings: \textit{(i)} Pairwise Network f-DP, which quantifies privacy leakage between user pairs under random-walk style decentralized communication; and \textit{(ii)} Secret-based f-local DP, enabling structured/correlated noise injection via shared secrets. They combine f-DP tools with Markov-chain concentration to capture privacy amplification effects due to sparse communication, local iterations, and correlated noise, and they empirically show improved utility (or tighter $(\epsilon,\delta)$ bounds at comparable utility) vs. Rényi DP-based accounting.

The practical value depends on whether the assumed communication mechanism (e.g., random-walk mixing) and any shared-secret structure match real deployments, and whether implementers can actually enforce the noise/secret assumptions consistently across peers.

\phantomsection \textbf{\textit{ERIS~\cite{fenoglio2025ERISEnhancingPrivacy}.}} \label{paragraph:eris}
ERIS is a serverless FL framework targeting privacy leakage and communication overhead for large models. It decentralizes aggregation via gradient partitioning across multiple client-side aggregators (so the update remains mathematically equivalent to FedAvg, while reducing the amount of visible model parameters to attackers) and pairs this with distributed shifted gradient compression to cut transmitted payloads while preserving convergence guarantees.

Privacy improves because no single aggregator observes a full update: leakage decreases as the number of aggregators grows, and experiments report reduced membership-inference and reconstruction effectiveness. Key limitations are that privacy weakens under aggregator collusion, and robustness to model/data poisoning is not a core part of the framework (left for future work).

\subsubsection{\underline{Security}}
\label{subsubsec:tdfl-security}
Without a trusted server, DFL must tolerate malicious peers that poison updates or collude to evade local defenses. Most TD-FL security mechanisms act locally at aggregation time: nodes infer neighbor reliability directly from received models (e.g., distance- or clustering-based consistency) and then downweight or discard suspicious updates (Trusted DFL~\cite{9700624}, BALANCE~\cite{fang2024byzantine-dfl}). Complementarily, BR-DFL~\cite{11230822} targets backdoor resilience by using knowledge distillation with adaptive teacher selection to suppress trigger-specific behavior. Finally, SelfishAttack~\cite{jia2025selfishattack} highlights a different threat model—strategic insiders optimizing for personal advantage rather than global failure—showing that robustness must also consider incentive-driven manipulation beyond classic poisoning.

\phantomsection \textbf{\textit{Trusted DFL~\cite{9700624}.}} \label{paragraph:trusted-dfl}
This work targets security, focusing on robustness to model poisoning. Each node maintains a trust score for its neighbors and uses it to downweight suspicious updates during the local aggregation step. Concretely, trust is inferred directly from received models using distance- and clustering-based consistency checks (e.g., comparing a neighbor’s update to other received updates, or verifying that the neighbor’s model moves closer over time), and then propagated/aggregated through the network via a trust-aggregation mechanism; the resulting trust values modulate how much each neighbor contributes to the node’s next iterate.

A key limitation is that the defense is heuristic and topology-/threshold-dependent: adversaries that adapt to the similarity tests, collude, or poison in a stealthier way can erode its effectiveness, and the paper mainly validates against a simple random noise model poisoning setup under simplifying assumptions (e.g., IID data).

\phantomsection \textbf{\textit{BALANCE~\cite{fang2024byzantine-dfl}.}} \label{paragraph:balance}
The authors study DFL in the presence of poisoning Byzantine attacks (data/model poisoning, targeted/untargeted, and adaptive attacks). They identify that existing DFL methods lack robustness against such attacks and propose a new algorithm that allows nodes to filter out possible malicious peers.

The proposed method filters neighbor updates before averaging: each client uses its own current model as a similarity reference to score received models and exclude likely-malicious ones. The paper proves convergence under poisoning in strongly convex and non-convex regimes, with rates matching strong baselines in attack-free cases, and reports extensive experiments where BALANCE improves robustness relative to prior DFL methods under poisoning.

However, BALANCE assumes the client’s own model remains a reliable similarity baseline; in extreme non-IID or in topologies where a client’s neighbors are mostly very different (or malicious), the filtering signal may be unreliable and information flow can be impeded. The method can also inherit topology sensitivity because “who you see” determines what you can accept, given that the authors assume a fixed topology.

\phantomsection \textbf{\textit{BR-DFL~\cite{11230822}}} \label{paragraph:brdfl}
BR-DFL is a backdoor-resilient DFL framework that does not rely on prior knowledge of attack types or the amount of malicious nodes. Unlike conventional defense approaches, in BR-DFL benign clients distill knowledge from peer models using their own clean local data. The key insight is that backdoored models behave normally on clean inputs; by performing knowledge distillation on trigger-free local data, benign clients transfer only clean predictive behavior, while backdoor behaviors are never activated and therefore not distilled. The paper also introduces a multi-armed bandit mechanism for adaptive teacher selection. Each benign node learns which peers provide the most distributionally aligned and informative knowledge, balancing exploration and exploitation. Empirically, BR-DFL achieves consistently low attack success rates across both data- and model-poisoning attacks and remains effective even under extreme malicious ratios ($> 50\%$). However, it introduces a non-trivial computational overhead given by the forward inference on the teacher models and the knowledge distillation process; moreover, the performance on attacks that target knowledge distillation~\cite{chen_taught_nodate} is unknown.

\phantomsection \textbf{\textit{SelfishAttack~\cite{jia2025selfishattack}.}} \label{paragraph:selfishattack}
The paper identifies a class of insider, strategic attacks in DFL where malicious clients do not necessarily want to collapse global performance; instead, they want their own models to end up better than others’ by manipulating what they send to non-selfish peers.
The authors formalize SelfishAttack as an optimization problem that balances attackers’ own utility and the degradation of non-selfish clients post-aggregation, and they derive closed-form optimal solutions for several aggregation rules. Empirically, they show larger attacker-favored accuracy gaps than standard poisoning baselines when adapted to the same objective.

The attack’s power is closely tied to the aggregation rules; for example, formal solutions cannot be derived for non-coordinate-wise aggregation rules. Another point is that peers are assumed to use the same aggregation strategy, while in real applications they may employ different methods. In this scenario, it is plausible to suppose that peers may also change aggregation rules over time and that this information is held secretly.

\subsubsection{\underline{Bandwidth utilization}} \label{subsubsec:tdfl-bandwidth-utilization}
In DFL, nodes exchange updates with peers rather than a server, so communication can dominate runtime—especially under dense topologies and heterogeneous, bandwidth-limited links. This makes bandwidth utilization a key scalability bottleneck.

The works in this subsection reduce communication by limiting peer interactions (gossip-style exchanges), selecting neighbors in a bandwidth-aware way, and shrinking payloads via compression or model segmentation (i.e., exchanging only parts of the model).

%

\phantomsection \textbf{\textit{Combo~\cite{hu2019decentralized}.}} \label{paragraph:combo}
Combo targets bandwidth utilization in DFL by reducing synchronization bottlenecks and exploiting parallel node-to-node transfers without sacrificing convergence. Methodologically, it uses a segmented gossip strategy: the model is split into equal-size parameter blocks, and each worker exchanges and aggregates only the corresponding segments with a small subset of peers; a “model replica” mechanism is introduced to improve information mixing across iterations. Experiments show faster training with limited accuracy loss under practical bandwidth/topology settings, and improved convergence behavior compared to FedAvg.

Its main limitations are that segmentation and replica management add coordination/engineering complexity, and performance depends on how segments are scheduled and mixed across the network (e.g., small peer degree or unfavorable connectivity can slow information propagation).

\phantomsection \textbf{\textit{GossipFL~\cite{9996127}.}} \label{paragraph:gossipfl}
GossipFL targets bandwidth-efficient decentralized training by reducing both the number of peer exchanges and the size of transmitted updates, while still aiming to preserve convergence. Methodologically, it combines (i) strong model sparsification so that each client exchanges a highly compressed update with only one peer per round, and (ii) a “gossip matrix” that assigns peer contacts based on link capacities to better utilize available network resources.

Its main limitations are architectural and assumption-driven: peer selection is coordinated by a central entity (making the system hybrid rather than fully decentralized), and the gains depend on having reliable bandwidth estimates and on how well the sparsification level matches the learning dynamics under non-IID data.

\subsubsection{\underline{Data heterogeneity}}
\label{subsubsec:data-heterogeneity}
Data heterogeneity is the non-IID regime where each node trains on a biased local distribution; in DFL, repeated neighbor-only aggregation can amplify this bias, making naive parameter averaging prone to client drift and slower convergence~\cite{zhao2018noniid}.

TD-FL works mainly mitigate this without sharing raw data by (i) replacing averaging with knowledge transfer, and/or (ii) adapting the communication graph so nodes aggregate mostly from similar peers. A complementary line uses probabilistic updates over arbitrary graphs, exchanging posterior information instead of full models

\phantomsection \textbf{\textit{Def-KT~\cite{9426904}.}} \label{paragraph:def-kt}
Def-KT targets data heterogeneity in DFL, specifically mitigating client drift that arises when peers fuse models by parameter averaging under non-IID data, which can slow/unstabilize convergence and degrade accuracy.
Methodologically, Def-KT performs mutual knowledge transfer: in each round, a subset of clients locally trains via SGD and sends its updated model to another randomly chosen client; the receiver then fuses the received and local models by making them match each other’s soft predictions on the receiver’s local data, and stores the fused result as its new local model. 
Limitations include added computation at the receiver, reliance on assumptions that simplify the system (a round structure where some clients only send and others only receive), and the lack of a communication-efficient variant or a full theoretical treatment in the paper (explicitly left as future work).

\phantomsection \textbf{\textit{P2P FL over graphs~\cite{9825726}.}} \label{paragraph:p2p-fl-over-graphs}
This work targets statistical heterogeneity in fully DFL by replacing point-model averaging with a Bayesian, variational formulation that is well-defined on arbitrary connected graphs. Each node maintains a posterior distribution over the shared model parameters and iterates two steps: \textit{(i)} a local variational update using its own labeled data batch, and \textit{(ii)} neighbor communication followed by a consensus step that merges information via log-posterior averaging with graph-dependent weights. The paper provides convergence guarantees in the realizable setting and characterizes convergence rates as a function of both network structure (via eigenvector centrality) and local data informativeness; it also reports experiments (including non-IID regimes and time-varying/asynchronous connectivity) showing improved behavior over FedAvg in harder, heterogeneous settings.

Main limitations are that the strongest theory relies on technical assumptions (e.g., connected/aperiodic graph, bounded likelihood ratios) and, in practice, performance depends on the chosen variational family and posterior approximation. The method also assumes the communication weights are available and stable enough to support the consensus mechanism.

\phantomsection \textbf{\textit{PANM~\cite{li2022panm}.}} \label{paragraph:panm}
PANM targets data heterogeneity in DFL by adaptively reshaping the communication graph so that each node mostly aggregates updates from “similar” peers, effectively inducing clustered training and improving personalization without assuming the number of clusters in advance. Methodologically, it is a two-stage neighbor-selection pipeline: first, each node builds a candidate neighbor pool using a gradient-based similarity score plus Monte Carlo exploration to reduce the risk of missing true peers; then it refines this pool with an EM-based Gaussian mixture model that separates likely in-cluster neighbors from outliers and augments/prunes the neighborhood accordingly. The resulting adaptive topology is used to run decentralized training where information flows primarily within clusters.

Its main limitations are that it relies on similarity estimates computed from local updates, which can be noisy under small local datasets, strong non-IID regimes, or rapidly drifting models, and it introduces extra coordination overhead (sampling rounds, EM fitting) before reaching a stable neighborhood. The Gaussian-mixture assumption may be a poor fit when similarities are multi-modal or adversarially manipulated, and poisoning/Sybil-style behavior can distort neighborhood discovery unless PANM is paired with explicit robustness mechanisms.

\phantomsection \textbf{\textit{DeSA~\cite{huang2024modelheterogeneity-dfl}.}} \label{paragraph:desa}
This work addresses the lack of a global model in DFL by synthesizing global anchor objects that encode shared distributional information, then using these anchors to regularize local training and enable mutual knowledge transfer across clients.

The method, DeSA, synthesizes anchors based on raw data distributions and introduces two training terms: a regularization loss that aligns each client’s latent embedding distribution with anchors, and a KD-based mechanism that lets clients learn from each other. The paper motivates the approach through domain-adaptation/KD perspectives and reports improved inter- and intra-domain accuracy across heterogeneous settings.

About the limitations, the anchor synthesis introduces computational and communication overhead, and the quality/representativeness of anchors becomes an additional dependency: poor anchors can limit transfer or bias representations.

\phantomsection \textbf{\textit{NTK-DFL~\cite{thompson2025ntkdfl}.}} \label{paragraph:ntk-dfl}
The paper proposes NTK-DFL, a DFL method that leverages Neural Tangent Kernel (NTK) to improve performance focusing on settings where client data distributions are heterogeneous. 

NTK-DFL uses NTK-based weight evolution together with model averaging across clients to exploit inter-client model variation, yielding faster convergence and higher accuracy compared to existing decentralized baselines, especially in heterogeneous settings. The method has been validated over multiple datasets, network topologies and heterogeneity levels. Nevertheless, NTK-based evolution requires computing and transmitting Jacobians in addition to weights, which comes with an additional overhead cost throughout the network.

\subsubsection{\underline{Theory and task extensions}}
\label{subsubsec:tdfl-theory-and-taskext}
In the following, we group TD-FL works whose main contribution is not a new communication or robustness mechanism per se, but rather \textit{(i)} theoretical foundations and statistical characterizations beyond standard convergence (e.g., asymptotic normality, uncertainty quantification, inference), or \textit{(ii)} task-level extensions that adapt DFL to primitives beyond gradient-based model training. These papers are complementary to challenge-specific methods: they clarify what guarantees DFL can provide and broaden the set of workloads that can be executed without a central coordinator.

\phantomsection \textbf{\textit{FDFL \cite{Lalitha2018FullyDF}.}} \label{paragraph:fdfl}
This approach eliminates the need for a centralized controller and instead focuses on collaborative learning through local information exchange. 
Specifically, clients adopt a Bayesian-like approach by introducing a belief over the model parameter space and using a distributed learning algorithm. Through information aggregation from neighbors, users update their beliefs to learn a model that best fits the observations across the entire network. The framework also allows users to sample points exclusively from smaller sub-spaces within the input space.

The paper provides high probability bounds on the worst-case probability of error across the network and explores appropriate approximations for applying the algorithm to learn DNN models. However, it is important to note that the work lacks experimental validation and is primarily a theoretical study.
Future studies could potentially complement this work with empirical experiments to validate its practical performance.

\phantomsection \textbf{\textit{BrainTorrent~\cite{DBLP:journals/corr/abs-1905-06731}.}} \label{paragraph:braintorrent}
BrainTorrent targets collaborative training across medical sites without a fixed coordinator, aiming to make decentralized coordination practical for a domain-specific task (medical image segmentation) and to keep training running under failures.

Methodologically, it uses a fully connected peer set with a lightweight “community” routine: nodes train locally, then an elected peer periodically polls others for their latest models and merges the received subset via sample-size–weighted averaging (and nodes keep a small cache of last-seen models/versions to manage asynchronous arrivals).

Limitations are mostly system and evaluation related: it assumes an expensive fully connected overlay and a single temporary “collector” whose failure can stall that round unless additional recovery logic is added; empirically, the validation is narrow and relies on task-specific metrics, so generality and scalability are not clearly established.

\phantomsection \textbf{\textit{Sharp Gaussian approximations~\cite{bonnerjee2025sharp}.}} \label{paragraph:sharpgaussianapproximations}
This paper studies statistical properties of local SGD in DFL beyond convergence, providing Gaussian approximation results for both the final iterate and the entire training trajectory, with the aim of enabling principled uncertainty quantification and inferential tools. The authors prove a Berry–Esseen theorem for final local-SGD iterates (supporting multiplier bootstrap procedures) and give time-uniform Gaussian approximations for the trajectory (enabling trajectory-level bootstrap tests). The validation is primarily simulation-based.

\phantomsection \textbf{\textit{Statistical inference for DFL~\cite{gu2024statisticalinferencefordfl}.}} \label{paragraph:statisticalinferencefordfl}
This work develops statistical inference theory for decentralized SGD under heterogeneous client distributions, analyzing mean squared error and consensus error, and establishing asymptotic normality results that enable confidence regions in DFL. The authors prove asymptotic normality for the Polyak–Ruppert averaged estimator in decentralized settings and show that statistical efficiency can impose stricter constraints on the number of clients than in some distributed M-estimation settings. They propose a one-step estimator that relaxes these constraints while achieving comparable efficiency, and they outline how to build confidence regions for practical inference. However, translating the conditions cleanly to nonconvex deep learning practice (and to dynamic network topologies) remains nontrivial.

\phantomsection \textbf{\textit{Excalibur~\cite{chai2024df-svd}.}} \label{paragraph:excalibur}
Excalibur is a system for performing federated SVD without a central server, designed to be communication-efficient and to avoid the heavy computational overhead of homomorphic encryption-based decentralized solutions. The goal is a practical decentralized pipeline for SVD that improves privacy relative to server-aided approaches by removing the external server, while maintaining strong efficiency. The paper explicitly positions itself against “server-aided federated SVD” and HE-based decentralized approaches. While lighter than HE-based implementations, the protection mechanism still introduces overhead, and the system’s applicability depends on whether the SVD workload and communication pattern align with the assumed decentralized execution model.

\subsection{Blockchained-based FL (BC-FL)}
\label{subsec:bc-fl}
The integration of blockchain and FL stems from their shared reliance on distributed coordination, but BC-FL should not be seen as a drop-in replacement for the server in standard FL. Rather, blockchain is used as a coordination and accountability layer that can reduce reliance on a single trusted orchestrator, provide tamper-evident logging, and support incentive settlement.

In BC-FL, participating nodes maintain a replicated ledger that records the progression of training rounds and related metadata. Due to size and throughput constraints, most systems avoid storing full model parameters on-chain; instead, they store hashes/commitments and pointers to off-chain storage (e.g., IPFS or edge/cloud repositories), which still makes model artifacts broadly retrievable and auditable. Training proceeds by having clients compute local updates and submit them (or references to them) to the blockchain workflow, where designated miners (or committees, in permissioned settings) verify admissibility according to the protocol (e.g., integrity checks, contribution rules, or robust screening), and then assemble a block that commits the accepted updates and the resulting aggregated model state. A consensus mechanism selects the next block proposer, after which the new block is propagated and appended, and clients fetch the latest global model (or its reference) to continue training.

Some BC-FL proposals additionally include reward mechanisms to encourage participation or discourage low-quality contributions; rewards may be tied to reported sample size, measured utility, reputation, or other protocol-specific signals, with the caveat that any self-reported signal must be made verifiable or it becomes a manipulation surface.

As for TD-FL, in the following section, we analyze the prominent BC-FL approaches proposed in the literature and categorize them based on the challenges that authors primarily aim to solve: \textit{fault-tolerance and scalability}, \textit{privacy and security}, \textit{incentive mechanisms}, and \textit{resource efficiency}.

\subsubsection{\underline{Fault-tolerance}} \label{subsubsec:bcfl-fault-toleranceandscalability}
In BC-FL, fault tolerance is pursued by replacing a single coordinator with ledger-driven coordination and replicated state, so that training can continue despite node failures and membership changes. Typical designs use (i) smart contracts to externalize orchestration (task control, membership, and key/state management) and (ii) structured topologies (often rings) to propagate models without a central hub, sometimes complemented by committee-based aggregation to provide redundancy and reduce the impact of unreliable peers. The works below follow this pattern: FED-BC~\cite{9182705} uses blockchain contracts to decentralize control and keep training alive under dropouts, while GFL~\cite{hu2021gfl} adds a trusted committee on top of a ring topology to stabilize coordination and provide a tamper-evident training log.

\phantomsection \textbf{\textit{FED-BC~\cite{9182705}.}} \label{paragraph:fed-bc}
FED-BC targets decentralization and fault tolerance in FL by removing a central server and avoiding a single point of failure, while also providing auditable coordination through a blockchain layer. The core idea is to organize participants in a logical ring for model propagation, while using smart contracts to handle control logic and key management (so no single node is responsible for orchestration). Each node trains locally, exchanges model updates along the ring, and relies on the ledger/contracts to record and coordinate the training process, aiming to keep the system running even if some nodes drop out.

Its main limitations are practical and assumption-driven: the approach inherits overhead from blockchain operations (latency, storage, and communication), and the “all nodes are equal” design still depends on correct smart-contract execution and on the ring not becoming fragmented under churn. The evaluation is mainly accuracy-oriented and does not deeply analyze adversarial behavior (e.g., poisoning) beyond the general robustness claims.

\phantomsection \textbf{\textit{GFL~\cite{hu2021gfl}.}} \label{paragraph:gfl}
GFL targets fault tolerance and coordination in BC-FL by avoiding reliance on a single aggregator, while also aiming to improve communication efficiency and provide tamper-evident logging. Methodologically, it organizes participants via consistent hashing into a ring topology and delegates aggregation to a committee of $m$ trusted nodes that synchronize models around the ring for $r$ steps. At each iteration, trusted nodes \textit{(i)} collect local models, \textit{(ii)} run a filtering/merging step based on knowledge distillation—each trusted node distills from selected peer models (e.g., those with smallest KL divergence) into its own model—and then (iii) compute the new global model via FedAvg over the $m$ trusted nodes. Blockchain and IPFS are used to store model references and coordinate tasks through smart contracts.

Limitations include strong trust assumptions (availability and honesty of the committee), extra overhead from blockchain/IPFS and multi-round synchronization, and an attack model that is only partially specified.

\subsubsection{\underline{Privacy}} \label{subsubsec:bcfl-privacy}
In BC-FL, the blockchain acts as a shared, replicated coordination layer. Depending on the design, the ledger can record training metadata and, in some cases, model artifacts themselves (global models and/or local updates) or—more commonly—cryptographic commitments and pointers to off-chain storage that make these artifacts broadly retrievable and auditable. This differs from TD-FL, where the visibility of updates is primarily determined by the communication topology (limited to a node’s neighborhood). As a result, BC-FL privacy mechanisms are motivated not only by gradient leakage, but also by the larger exposure surface induced by ledger-based coordination and validation. Notably, many BC-FL systems avoid storing full model parameters on-chain due to size and throughput constraints; instead they log hashes on-chain while keeping the actual updates/models off-chain, which still enables network-wide access for validation and audit.
Accordingly, the BC-FL works in this subsection enhance privacy requirements by employing DP~\cite{9170559}, encrypted gradients~\cite{8894364}, and secure aggregation~\cite{9292450}.

\phantomsection \textbf{\textit{Privacy-preserving FL for IoT devices~\cite{9170559}.}} \label{paragraph:privacy-preservingflforiotdevices}
This work targets privacy-aware BC-FL for smart-home IoT, aiming to let manufacturers learn from customer data without collecting it centrally, while discouraging malicious or low-quality participants. The system uses a blockchain layer for coordination and logging: customers train locally (optionally with MEC support) and submit updates, while selected miners aggregate them into a global model. Privacy is strengthened via differential privacy applied to client updates, and the paper proposes a DP-friendly normalization technique to mitigate accuracy degradation under noise. To improve integrity, it combines an incentive/reputation mechanism with Byzantine-robust aggregation (Multi-Krum) to reduce the impact of unreliable clients.

Key limitations are that robustness hinges on how miners are selected and behave (collusion/compromise can undermine integrity), and that Multi-Krum plus reputation are not tailored to stealthy poisoning/backdoor behavior and may be manipulated strategically. Evaluation is mostly benchmark-based, leaving realistic smart-home conditions (strong non-IID streams, churn, real appliance datasets) less explored.

\phantomsection \textbf{\textit{DeepChain~\cite{8894364}.}} \label{paragraph:deepchain}
DeepChain targets privacy and correctness in BC-FL by protecting local training updates while making the aggregation process publicly auditable and economically fair. The core idea is to treat training updates as blockchain transactions: participants encrypt their intermediate gradients and submit them on-chain; designated workers collect and process these transactions to update the global model, and the updated model state is recorded on the ledger. Privacy is enforced via threshold Paillier encryption so that gradients remain confidential while still supporting encrypted aggregation. Auditability is achieved by attaching verifiable proofs to transactions, allowing others to check that collection and parameter updates were performed correctly. The framework also includes an incentive mechanism with rewards and penalties (e.g., timeout and monetary punishment) to discourage misbehavior. Limitations include substantial cryptographic/protocol complexity and reliance on correct incentive calibration and honest-majority assumptions for the chain’s consensus/workers.

\phantomsection \textbf{\textit{Biscotti~\cite{9292450}.}} \label{paragraph:biscotti}
Biscotti targets privacy and robustness by enabling decentralized training without revealing individual client updates, while limiting the impact of malicious participants. The system replaces a central server with a blockchain that logs training rounds and coordinates three rotating committees: proposers (collect updates), validators (check update validity), and aggregators (compute the global update). Privacy is provided through secure aggregation based on secret sharing so that only an aggregated update is revealed, not any single client contribution. Robustness is strengthened via a validation step that filters suspicious updates using anomaly detection before aggregation, and the committees are selected via a stake-based/PoS-like mechanism to reduce Sybil influence. Limitations include extra protocol complexity (committee formation, validation, secret sharing) and reliance on assumptions about committee honesty and sufficient benign participation; if attackers control committee selection or adapt to the validation filter, robustness can degrade.

\subsubsection{\underline{Security}} \label{subsubsec:bcfl-security}
Security in BC-FL primarily concerns integrity of the learning process under malicious or unreliable participants: poisoning (model/data), low-quality updates, and manipulation of aggregation/validation. The surveyed works commonly enforce integrity by (i) using consensus to validate and finalize updates (permissioned PBFT-style validation in Trustworthy FL~\cite{9866512}, HotStuff-style SMR in PIRATE~\cite{zhou2019pirate}), (ii) filtering or scoring updates before they influence aggregation, via robust aggregation rules and anomaly/reliability assessment (Multi-Krum and reputation in~\cite{9170559}, anomaly-based scoring/committee aggregation in PIRATE~\cite{zhou2019pirate}), and (iii) designing consensus incentives around model utility so that “useful” updates are favored during block production (proof-of-accuracy in BytoChain~\cite{9382023}). While these mechanisms can raise the bar for straightforward attacks, they typically shift the problem to the reliability of validation signals (accuracy-based thresholds, anomaly detectors, committee honesty) and to the practicality of running repeated verification rounds at scale.

\phantomsection \textbf{\textit{BytoChain~\cite{9382023}.}} \label{paragraph:bytochain}
BytoChain targets robustness to Byzantine (poisoning) behavior in BC-FL at the edge, while keeping verification delay manageable. Its core idea is “dual verification”: verifiers first screen submitted local models in parallel using a small validation subset and discard models whose accuracy drop exceeds an Accuracy Oscillation Threshold (AOT); miners then build a block by selecting enough verified transactions and proposing the averaged model with the best validation accuracy, subject to AOT. Consensus is Proof-of-Accuracy (PoA): other miners accept a block only if (i) the AOT constraint holds, (ii) Accuracy Deviation Threshold (ADT) bounds cross-miner accuracy disagreement, and (iii) a homomorphic-hash constraint proves the global model is the average of the referenced local models without downloading all parameters (stored in IPFS, while hashes go on-chain). Limitations include reliance on validator/miner-held validation subsets, sensitivity to AOT/ADT tuning, and incomplete coverage of stealthy attacks (the paper flags privacy/confidentiality and backdoors as open issues).

\phantomsection \textbf{\textit{PIRATE~\cite{zhou2019pirate}.}} \label{paragraph:pirate}
PIRATE targets Byzantine-resilient decentralized training, focusing on protecting gradients/model parameters and keeping aggregation reliable under malicious behavior. Methodologically, it combines (i) a sharded blockchain that runs a BFT state-machine replication protocol (\textit{hotstuff}~\cite{ittai2018hotstuff}) to agree on an append-only log of training “commands” and partial aggregation results, and (ii) a committee-based workflow where nodes are randomly split into multiple committees that perform partial aggregations in parallel to reduce load. To limit poisoning, each committee applies a reliability assessment step that scores incoming gradients with a pre-trained anomaly detector and downweights/excludes suspicious updates before committing results.

Limitations are mainly evaluation and assumptions: the prototype is compared to a baseline largely under no-adversary settings (communication/storage and iteration time), while the robustness claims rely on how well the anomaly detector generalizes and on stable committee formation and BFT assumptions (e.g., honest-majority per committee and network conditions).

\phantomsection \textbf{\textit{Trustworthy FL~\cite{9866512}.}} \label{paragraph:trustworthyfl}
This work targets trustworthy BC-FL at the wireless edge by protecting the integrity of global updates while keeping the coordination layer energy-efficient. The architecture combines a permissioned blockchain with wireless FL: global model updates are validated through Practical Byzantine Fault Tolerance (PBFT), which avoids PoW’s energy cost but assumes a permissioned committee and tolerates up to one-third malicious edge servers. The paper details the PBFT-based training pipeline and highlights that consensus induces non-negligible training latency due to repeated cross-validation/verification rounds among edge servers.

To reduce this latency, the authors formulate a network optimization problem that jointly allocates bandwidth and transmission power to minimize long-term average training delay across rounds. They cast it as a Markov decision process and solve it with a DRL method (TD3), aiming for good performance with low online complexity. For robustness against poisoning on the client side, the framework uses Multi-Krum as the aggregation rule.

Limitations mainly come from the permissioned setting and PBFT overhead: scalability depends on the size/connectivity of the edge-server committee, and the empirical validation is mostly benchmark/simulation-driven with simplified adversarial conditions.

\subsubsection{\underline{Incentive mechanisms}}
\label{bcfl-subsubsec:incentivemechanisms}
In BC-FL, incentives are often coupled with the coordination layer: the ledger records who contributed what, while rewards are used to (i) attract participants with valuable data or resources and (ii) discourage free-riding and low-quality updates. The two works below represent two common directions: BlockFL~\cite{8733825} rewards devices proportionally to their training sample sizes and analyzes how consensus latency affects training delay, whereas FedCoin~\cite{liu2020fedcoin} uses a payment layer where rewards are computed from each client’s measured contribution to the learned model.

\phantomsection \textbf{\textit{BlockFL~\cite{8733825}.}} \label{paragraph:blockfl}
BlockFL targets \textit{(i)} removing the single point of failure of a server and \textit{(ii)} incentivizing participation by paying devices that contribute more training data. It replaces the server with a blockchain network where devices upload local model updates to miners; miners verify updates, run PoW, and append a block that records the verified updates. Each device then computes the global update locally from the new block, so a miner/device failure does not halt training. Incentives are proportional to the local dataset used for training. The paper’s main limitation is that the truthfulness of the dataset size is assessed heuristically.

\phantomsection \textbf{\textit{FedCoin~\cite{liu2020fedcoin}.}} \label{paragraph:fedcoin}
FedCoin targets fair incentive allocation in FL by introducing Proof-of-Shapley-Value (PoSV): miners compute (or approximate) a Shapley-value vector that estimates each client’s marginal contribution to the final model, and the resulting reward split is recorded immutably as payments on-chain. The architecture separates an FL network from a P2P blockchain network that handles accounting and nonrepudiation. Limitations are that Shapley-value computation is expensive and typically requires approximations and repeated utility evaluations; rewards can also be sensitive to the chosen valuation protocol and to noisy/non-IID evaluation data.

\subsubsection{\underline{Resource efficiency}}
\label{subsubsec:resourceefficiency}
BC-FL introduces extra system costs (e.g., block production/verification delay, transaction backlogs, and additional compute/communication at edge nodes). As a result, resource-efficient BC-FL works typically co-design the learning loop with the ledger protocol, tuning when devices train/upload and how fast blocks are generated so that learning progress is not throttled by the chain. The two representative directions are: \textit{(i)} asynchronous FL with block-rate and participation control to avoid congestion and reduce delay/energy~\cite{9399813}, and \textit{(ii)} joint modeling of training, communication, and mining, with optimized role scheduling and resource allocation to minimize end-to-end training time under device constraints~\cite{9664296}.

\phantomsection \textbf{\textit{BAFL~\cite{9399813}.}} \label{paragraph:bafl}
It targets efficiency in BC-FL by reducing training delay and device energy consumption under asynchronous participation. It allows devices to upload local models without strict round synchronization and uses the blockchain for logging and coordination. The core methodology is a workflow control layer that \textit{(i)} adjusts the block generation rate to balance confirmation latency against congestion, and \textit{(ii)} adapts devices’ local training/upload timing to limit transaction overload. It also records participation-related scores (used as a notion of contribution/trust) on-chain and formulates a multi-objective optimization (energy--delay trade-off) to choose operating points. Limitations include that performance depends on accurate tuning of arrival rates and block parameters, and the analysis primarily focuses on efficiency rather than adversarial settings beyond a simple random-noise attack.

\phantomsection \textbf{\textit{BLADE-FL~\cite{9664296}.}} \label{paragraph:blade-fl}
It targets resource efficiency and robustness to single-point failures by integrating training and blockchain operations at the client side. Each client trains locally, broadcasts its update as a transaction, and participates in block creation/validation so that aggregation is driven by the ledger rather than a central server. Methodologically, the paper models the end-to-end learning cycle (local compute, transmission, block generation and validation) and derives learning-performance characterizations to guide system design; it then optimizes system parameters (e.g., the number of integrated rounds and resource allocation variables) to reduce long-term training latency while keeping learning progress stable. Limitations include reliance on simplifying assumptions for tractable analysis (e.g., homogeneous device capabilities and idealized communication/attack models).

\section{Lessons Learned \& Future Directions}
\label{sec:future}
In this section, we summarize the main lessons learned from the surveyed literature and outline future research directions for DFL, following the taxonomy in Section~\ref{sec:decentralized-fl} and emphasizing cross-cutting gaps across security, privacy, incentives, communication efficiency, and realistic deployment constraints.

\subsection{Security against Adversarial Attacks}
While in standard FL the coordinator typically does not hold a validation dataset, prior work has largely pursued robust aggregation rules~\cite{yin2018fedmedian, xie2018trimmedmean, blanchard2017krum, mhamdi2018bulyan, ma2022shieldfl} and filtering strategies~\cite{gabrielli2025flanders, zhang2022fldetector} that rely on statistical, geometric, or temporal consistency signals to assess incoming updates. In DFL, by contrast, each peer naturally has access to local labeled data that can act as a validation set. This opens the door to validation-based sanity checks of neighbor models prior to aggregation; a design explicitly explored in BytoChain and PANM, where nodes use a small held-out set not only to monitor their own training but also to score received peer updates before merging. Still, the advantage of per-node validation over distance- or statistical-based defenses is not guaranteed: local validation sets can be small and biased, and they are poorly suited to detecting backdoors because triggers are unlikely to appear in the held-out data. As a result, validation is best viewed as one stage in a defense pipeline (e.g., pre-filtering before robust aggregation, complemented by backdoor-oriented tests or mitigation).

This pipeline perspective is particularly important because importing server-based robust aggregators into DFL “as is”~\cite{9182705, 9170559, 9292450, 9866512} can create a false sense of security. Many aggregation defenses assume that, within the set of updates being aggregated, the malicious fraction is bounded (often $<\frac{1}{2}$, sometimes $<\frac{1}{3}$ depending on the rule and threat model). While such assumptions can be reasonable when a server samples broadly from a large population, they may fail locally in DFL: \textit{a node can be surrounded by adversaries within its neighborhood even if the global network is largely benign}. Consequently, local validation can serve as a first-line filter to reduce obviously harmful updates, after which robust aggregation (and, when relevant, topology- and backdoor-aware checks) can be applied to the remaining neighbor models.

A recurring lesson from the reviewed BC-FL literature is that the blockchain is often used primarily as a logging substrate that partially replaces the server’s role, rather than as a security panacea. Moreover, many designs shift trust to miners or validators, \textit{creating a new hierarchy resembling centralized FL}. Other than that, BC-FL methods may expand the attack surface by introducing consensus- and smart-contract-related failure modes on top of the learning-layer threats.

In general, DFL enlarges the adversarial surface compared to server-based FL: updates are exchanged over P2P links, the topology influences exposure and attack propagation, and the absence of a single coordinator complicates monitoring and mitigation. In BC-FL, consensus and trust management can incentivize correct behavior and help identify falsified updates, but the attack landscape is evolving, and current designs still face open threats. Although several surveyed systems demonstrate robustness against basic adversarial behaviors, the literature still lacks a systematic security evaluation for DFL under realistic and adaptive attacker models and under diverse topologies. A key future direction is to consolidate threat models and benchmarks that explicitly capture \textit{(i)} topology-aware poisoning/backdoor attacks, \textit{(ii)} Sybil and collusion strategies in decentralized settings, and \textit{(iii)} the interaction between security measures and non-IID heterogeneity. In parallel, defenses should move beyond ``single-mechanism'' approaches by combining robust aggregation/filtering, trust/reputation signals, and protocol-level constraints (e.g., neighbor selection rules) in a way that is still compatible with privacy and communication limits, while fostering cooperation.

\subsection{Strategic and Utility-Seeking Insider Threats}
Most security work in DFL is framed around disruption-oriented adversaries and defenses that aim to preserve convergence or accuracy. However, decentralization naturally enables strategic behaviors that optimize a participant’s own utility rather than purely degrading global performance. This is further confounded by statistical heterogeneity, since benign updates can already be highly variable and neighborhood-specific, making strategic deviations harder to distinguish from non-IID behavior using proxy metrics alone. 
Competitive advantage attacks~\cite{jia2025selfishattack} exemplify this threat model and highlight that standard Byzantine-robustness is an incomplete evaluation lens. A key future direction is to develop security benchmarks and defenses explicitly designed for utility-seeking adversaries, including collusion, and to study protocol-level countermeasures that combine robust filtering/aggregation with incentive-aware rules that discourage strategic deviations.

\subsection{Privacy in Adversarial Settings}
Privacy risks in DFL are generally amplified because local updates can be observed by multiple peers over time (and potentially by colluding neighborhoods), rather than by a single coordinator. \textit{In BC-FL, the exposure surface can further increase because ledger-driven coordination makes updates/models broadly retrievable, whereas in TD-FL visibility is largely constrained by the communication topology}. BC-FL also often introduces a miner/validator hierarchy; if these parties are treated as more trusted than peers, \textit{the trust model drifts toward a setting resembling the centralized FL setting} where trust is assumed.

Recent works motivate privacy mechanisms tailored to decentralization, including \textit{(i)} privacy accounting schemes explicitly designed for DFL communication patterns (e.g., f-differential privacy~\cite{li2025mitigating}), and \textit{(ii)} privacy-by-design approaches that reduce leakage through protocol structure and partial observability (e.g., ERIS~\cite{fenoglio2025ERISEnhancingPrivacy}). A major open problem is to unify these perspectives into threat models and privacy metrics that reflect realistic DFL deployments, including compromised subsets of peers, collusion, time-varying topologies, and repeated exposure across rounds. Another promising direction is to develop standardized privacy benchmarks for DFL and, thereafter, jointly measure privacy leakage with accuracy, and communication cost, since recent approaches explicitly target these trade-offs simultaneously.

\subsection{Incentive Mechanisms}
Incentives are a core bottleneck for DFL because participation is voluntary and costly: nodes should contribute computation despite free-riding opportunities~\cite{Lewis2023AttacksAF, fraboni2021freerider}, privacy constraints, and the difficulty of evaluating contribution fairly. This challenge is only partially addressed in current work: in TD-FL, incentive mechanisms are essentially absent in our corpus, with IPLS~\cite{9472790} only briefly mentioning the integration of a cryptocurrency-like. In BC-FL, rewards exist but are often tied to data quantity~\cite{8733825} or measured accuracy~\cite{liu2020fedcoin}, which can discourage weaker nodes and does not directly solve “who should be rewarded for improving the model.” 

Two structural issues make incentives in FL harder than in many P2P systems: nodes do not share strategic information due to privacy concerns, and contribution-to-accuracy is difficult to estimate and highly nonlinear. In DFL, \textit{incentives also interact with topology: who exchanges with whom affects both measurability of contribution and manipulability of reward signals.} A concrete future direction is to develop incentive mechanisms that are \textit{(i)} compatible with privacy (minimal disclosure), \textit{(ii)} robust to manipulation (e.g., gaming reputation or falsifying contribution evidence), and \textit{(iii)} topology-aware. Promising starting points include adapting known P2P incentive families (monetary, auction-based, reputation-based, service-based) and leveraging game-theoretic tools, but grounding these designs in DFL-specific constraints and evaluation metrics.  Finally, incentives should be studied jointly with security: strategic participants may deviate not only to free-ride but also to gain advantage, so incentive protocol design becomes a security primitive rather than a purely economic add-on.

\subsection{Fault Tolerance in Realistic Decentralized Applications}
Fault tolerance is one of the main motivations for moving away from a centralized coordinator. In BC-FL, replicated ledger state can remove the single point of failure associated with a coordinator by externalizing global model progression into a shared log; however, robustness then depends on the liveness and partition tolerance of the consensus layer and on how model availability is ensured (on-chain vs off-chain).

Moreover, many DFL protocols are evaluated under relatively static assumptions (fixed participant sets, stable connectivity, synchronous rounds). A practical research gap is to characterize and improve DFL under churn, intermittent links, and asynchronous operation, especially when these effects interact with security and incentives. Developing reference simulation settings and datasets for dynamic environments, together with theory that captures time-varying graphs, would make empirical results more comparable and would push the literature closer to real-world applications. Another concrete gap is recovery under network partitions: when the graph splits, how should sub-networks continue training, and how (and whether) should models be reconciled when connectivity returns without amplifying poisoning or drift?

\subsection{Bandwidth Utilization and Communication Scalability}
Communication is the dominant scalability bottleneck in DFL because each peer exchanges updates over bandwidth-limited and heterogeneous links, and the topology can induce quadratic message growth in dense graphs. A key lesson from the surveyed TD-FL works is that communication savings come from three levers that are often coupled: exploiting link heterogeneity through neighbor selection, and reducing payload size via compression or by splitting exchanged models. However, these mechanisms introduce new failure modes: sparse communication can slow mixing and exacerbate non-IID drift, and aggressive compression can interact poorly with robustness defenses that rely on geometric distances.

Promising directions include topology-adaptive protocols that explicitly trade mixing speed for bandwidth, and co-designing compression with adversarial robustness mechanisms so that bandwidth savings do not weaken poisoning detection.

\subsection{Rethinking the notion of a global model in TD-FL}
Decentralization changes what ‘global’ means. In TD-FL, the training state is locally maintained and propagated along the topology; a single global model is only an abstraction when the graph is sufficiently connected and mixes fast enough for consensus to form. \textit{In realistic settings with churn, clustered communities, or intermittent connectivity, TD-FL may instead converge to multiple cluster-specific solutions}. In contrast, BC-FL designs often expose a network-wide shared state, making the global model closer to its meaning in centralized FL. 

On a related note, a recurring lesson in the TD-FL literature is that non-IID is not only a statistical issue, but also a topology one: who exchanges with whom shapes which distributions are mixed and, consequently, what notion of “global” model is achievable. The TD-FL works surveyed address heterogeneity without sharing raw data mainly by replacing parameter averaging with knowledge transfer, by clustering/neighbor matching so that aggregation happens among similar peers, or by exchanging probabilistic information rather than full models. These approaches implicitly move DFL along a spectrum from global consensus toward clustered or personalized solutions.

Future work should make this objective explicit: are we optimizing a single network-wide model, multiple cluster-level models, or fully personalized models? This calls for evaluation protocols that report not only average accuracy, but also cluster-wise behavior, fairness across communities, sensitivity to graph partitions, and robustness when clusters are adversarially manipulated (e.g., Sybils steering neighbor matching).
\section{Conclusion}
\label{sec:conclusion}
Federated learning (FL) has become a widely adopted paradigm for training machine learning models without direct data pooling. However, standard FL typically relies on a central coordinator, which introduces a single point of failure and a strong trust and security dependency on the orchestration infrastructure. Decentralized federated learning (DFL) removes this coordinator and enables peer-to-peer cooperation, but this shift makes information flow topology-dependent and introduces new challenges in security, privacy, and reliability.

In this survey, we reviewed the literature on DFL and organized existing approaches into two families—traditional distributed computing (TD-FL) and blockchain-based (BC-FL)—then structured them through a unified, challenge-oriented taxonomy. This view clarifies which bottlenecks each line of work targets and highlights recurring gaps across the field.

Several open problems emerge. First, security needs evaluation under decentralized threat models that account for topology-aware poisoning/backdoors, collusion, Sybil behavior, and utility-seeking insiders. Second, privacy guarantees should reflect decentralized exposure: in TD-FL, visibility is largely neighborhood-driven, while in BC-FL the shared ledger and its off-chain retrieval mechanisms can widen access, motivating privacy mechanisms that go beyond server-based assumptions. Third, fault tolerance remains underexplored under churn, intermittent connectivity, heterogeneous resources, and asynchronous operation, where even the notion of a single global model may be task- and topology-dependent.

We hope this survey serves as a reference for researchers and practitioners by mapping the DFL design space with a common taxonomy and by identifying the gaps that must be addressed to move from conceptual protocols to robust deployments.

\bibliographystyle{IEEEtran}
\bibliography{references}

\end{document}